\documentclass[letterpaper]{article} 
\usepackage{aaai2027}  
\usepackage[hyphens]{url}  
\usepackage{graphicx} 
\usepackage{natbib}  
\usepackage{caption} 
\usepackage{algorithm}
\usepackage{algorithmic}

\usepackage{amsmath}
\usepackage{amssymb}
\usepackage{mathtools}
\usepackage{amsthm}
\usepackage{booktabs}
\usepackage{siunitx}      
\usepackage[table]{xcolor}  
\usepackage{etoolbox}     
\usepackage{multirow}
\usepackage{amssymb}
\usepackage{marvosym}

\usepackage{newfloat}
\usepackage{listings}
\DeclareCaptionStyle{ruled}{labelfont=normalfont,labelsep=colon,strut=off} 
\floatstyle{ruled}
\newfloat{listing}{tb}{lst}{}
\floatname{listing}{Listing}

\usepackage{booktabs}

\title{Text-Driven 3D Indoor Scene Synthesis in Non-Manhattan Environments}
\author{
 \textbf{Xianhui Meng\textsuperscript{\rm 1,\rm 2}},
 \textbf{Zirui Song\textsuperscript{\rm 2,\rm 3}},
 \textbf{Yuchen Zhang\textsuperscript{\rm 2,\rm 4}},
 \textbf{Li Zhang\textsuperscript{\rm 1}},
\\
 \textbf{Yongxuan Lv\textsuperscript{\rm 1}},
 \textbf{Xiuying Chen\textsuperscript{\rm 3}},
 \textbf{Kun Wang\textsuperscript{\rm 5}},
 \textbf{Yan Luo\textsuperscript{\rm 6}},
 \textbf{Kai Chen\textsuperscript{\rm 7}},
 \textbf{Hangjun Ye \textsuperscript{\rm 2}},
\\
 \textbf{Long Chen\textsuperscript{\rm 2}},
 \textbf{Jun Liu\textsuperscript{\rm 1}}\corresponding,
 \textbf{Xiaoshuai Hao\textsuperscript{\rm 2}}\corresponding,
}

\affiliations {
 \textsuperscript{\rm 1}University of Science and Technology of China
,
 \textsuperscript{\rm 2}Xiaomi EV, \\
 \textsuperscript{\rm 3}Mohamed bin Zayed University of Artificial Intelligence,
 \textsuperscript{\rm 4}Georgia Institute of Technology,\\
 \textsuperscript{\rm 5}Nanyang Technological University,
 \textsuperscript{\rm 6}Massachusetts Institute of Technology,\\
 \textsuperscript{\rm 7}The Hong Kong University of Science and Technology,
\\
 junliu@ustc.edu.cn, haoxiaoshuai@xiaomi.com
 
}

\begin{document}

\maketitle

\begin{abstract}
Large Language Models (LLMs) have demonstrated remarkable capabilities in 3D indoor synthesis for Manhattan environments. However, existing methods often fail to capture plausible object layout patterns in non-Manhattan settings, primarily because they struggle to model non-orthogonal spatial relationships, leading to high geometric violations and low physical fidelity. To address this challenge, we propose \textbf{\textit{SPG-Layout}}, a novel text-driven framework designed to generate physically plausible indoor scenes within complex non-Manhattan environments. Specifically, we first utilize \textit{statistical priors} of object distributions to guide the training process, enhancing environmental understanding and fidelity. Furthermore, mirroring human design workflows, we adopt a \textit{hierarchical layout strategy} that prioritizes the placement of large objects, thereby substantially minimizing layout violations. By synergizing these components, \textbf{\textit{SPG-Layout}} achieves a balanced optimization of semantic realism and physical plausibility. To evaluate performance in these complex settings, we constructed a new benchmark comprising 500 diverse non-Manhattan environments. Extensive experiments demonstrate that \textbf{\textit{SPG-Layout}} consistently and significantly outperforms existing methods across both Manhattan and non-Manhattan environments. The code will be publicly released.
\end{abstract}


\section{Introduction}
\begin{figure}
    \centering
    \includegraphics[width=0.99\linewidth]{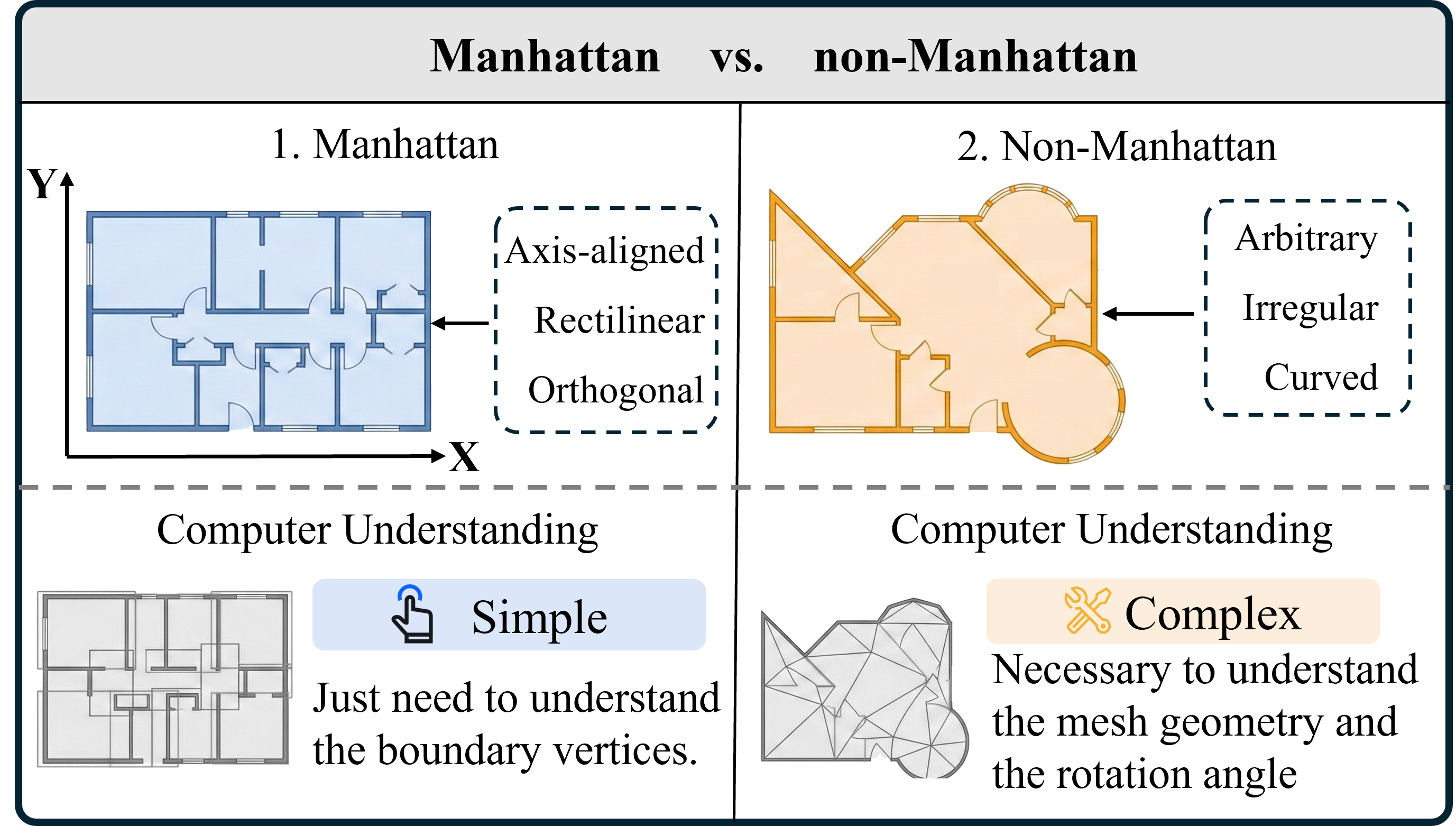}
    \caption{
   \textbf{Manhattan \textit{vs.} Non-Manhattan Environments.} Manhattan environments (left) are simple, characterized by orthogonal structures. In contrast, non-Manhattan environments (right) are difficult to understand due to their arbitrary, irregular, and curved geometries, requiring precise reasoning about mesh details and rotation angles. }
    \label{fig:1}
    \vspace{-1em}
\end{figure}

Recent advancements in Large Language Models (LLMs) have revolutionized text-driven 3D scene synthesis, enabling users to generate complex environments using natural language~\cite{bucher2025respace,yang2024holodeck,tang2024diffuscene,hu2024mixed,maillard2024debara,sun2024layoutvlm,feng2023layoutgpt,hao2025mimo}. However, a critical disparity remains between academic benchmarks and real-world architectural complexity. The vast majority of existing methods predominantly focus on Manhattan environments, characterized by simplified layouts with orthogonal walls and axis-aligned structures. While the \textit{Manhattan Assumption} facilitates grid-based generation, it significantly limits applicability in practical scenarios. In real-world indoor scenes, non-Manhattan topologies are not merely edge cases but are, in fact, prevalent. Contemporary architecture frequently employs complex geometric primitives, characterized by arbitrary turning angles, curvilinear boundaries, and irregular polygons, as illustrated in Fig.~\ref{fig:1}.

However, extending current generative paradigms to non-Manhattan settings presents non-trivial challenges. First, existing approaches~\cite{bucher2025respace,song2025hazards,yang2024holodeck,wang2023robogen,zhai2024echoscene,lin2024instructscene,sun2024forest2seq,feng2023layoutgpt,feng2025casagpt} learned distributions implicitly assume axis-aligned boundaries, so high-likelihood placements in Manhattan rooms may become physically infeasible once walls turn obliquely. Second, existing methods~\cite{ ccelen2025design, fu2025anyhome, hu2024scenecraft,  littlefair2025flairgpt, yang2024llplace, wang2024chat2layout, aguina2024open, xia2024scenegenagent} suffer from a semantic-spatial misalignment. While LLMs~\cite{yang2024qwen2,shao2024deepseekmath} excels at decomposing abstract instructions into object lists, it lacks the intrinsic geometric intuition to model the strong conditional spatial dependencies between objects (\textit{e.g.}, the canonical relative pose between a bed and a nightstand). In non-Manhattan environments, where spatial relations cannot be implicitly learned via axis-aligned patterns, this deficiency leads to layouts that are physically valid but functionally incoherent.
To bridge these gaps, we present \textbf{\textit{SPG-Layout}}, a novel text-driven framework tailored for generating physically plausible indoor scenes within complex non-Manhattan environments.

To tackle the issue of spatially incoherent layouts, we propose a two-stage training paradigm empowered by \textit{Spatial Prior Guidance}. Specifically, we align the LLM with structured scene representations and subsequently optimize the model using reinforcement learning. 
To ensure generation quality, we design verifiable rewards derived from the statistical distribution of non-Manhattan environments.
Furthermore, to mitigate physical collisions, we introduce a \textit{Hierarchical Layout Strategy} inspired by human interior design cognition.
By adopting a coarse-to-fine workflow that prioritizes large-scale furniture before allocating smaller items to the remaining space, we effectively mitigate spatial conflicts and ensure physical plausibility, even in constrained geometries.
Validating performance in non-Manhattan settings remains challenging due to data scarcity. 
To address this, we curated a novel dataset of 500 high-quality non-Manhattan scenes. We extracted irregular boundaries from publicly available professional floor plans and generated initial 3D layouts, which were subsequently polished via manual refinement to ensure quality. This dataset is then used to benchmark model performance in non-Manhattan environments. For completeness, we also conducted extensive experiments on standard Manhattan datasets. Empirical results demonstrate that our proposed \textbf{\textit{SPG-Layout}} significantly outperforms the existing state-of-the-art baselines in both Manhattan and non-Manhattan environments, establishing itself as a versatile and robust framework for text-driven 3D indoor scene synthesis. 

Our contributions are threefold:
\textbf{\textit{First}}, we propose \textbf{\textit{SPG-Layout}}, the first text-driven framework specifically optimized for non-Manhattan indoor environments. By breaking the ``Manhattan Assumption", our approach enables physically plausible and semantically coherent 3D scene generation in complex architectural spaces with irregular boundaries and arbitrary angles.
\textbf{\textit{Second}}, we introduce a dual-optimization strategy that mirrors human design cognition: \textit{(i)} \textit{Spatial Prior Guidance (SPG)}, which integrates statistical distributions of object-boundary and object-object relationships into a reinforcement learning reward ; and \textit{(ii)} a \textit{Hierarchical Layout Strategy (HLS)}, which prioritizes large-scale furniture to resolve spatial fragmentation and minimize physical collisions.
\textbf{\textit{Third}}, we propose the benchmark consisting of 500 diverse non-Manhattan environments to bridge the gap in existing datasets. Extensive evaluations demonstrate that \textbf{\textit{SPG-Layout}} significantly outperforms current methods in both layout fidelity and violation metrics across Manhattan and non-Manhattan settings, establishing a robust baseline for future research.

\section{Related Work}

\subsection{Indoor Scene Layout Generation}

Early indoor scene layout generation relied on heuristic rules, procedural modeling, and optimization-based techniques \cite{qi2018,weiss2018,fisher2015,purkait2020}, where handcrafted spatial constraints and expert knowledge were designed to produce functional and physically plausible layouts. With the emergence of deep learning, data-driven approaches gradually replaced manually designed rules. Representative methods include CNN-based scene priors \cite{wang2018deep}, autoregressive transformer models \cite{ritchie2019fast,wang2021sceneformer,paschalidou2021atiss,sun2024forest2seq}, and more recently diffusion-based approaches \cite{tang2024diffuscene,hu2024mixed,wei2023lego,maillard2024debara,yang2024physcene}. By learning object distributions from large-scale indoor scene datasets, these methods substantially improve layout realism and generation quality. Nevertheless, they primarily rely on learned spatial priors and predefined conditioning signals, making it difficult to flexibly synthesize scenes according to diverse user requirements.

\subsection{Language-Guided Indoor Scene Generation}

Recent advances in large language models (LLMs) have enabled indoor scene synthesis directly from natural language descriptions. Early work such as CLIP-Layout \cite{liu2023cliplayout} explored language-conditioned layout generation using vision-language representations. Building upon instruction-following LLMs, subsequent methods—including LayoutGPT \cite{feng2023layoutgpt}, LayoutVLM \cite{sun2024layoutvlm}, I-Design \cite{celen2024idesign}, Holodeck \cite{yang2024holodeck}, LLPlace \cite{yang2024llplace}, InstructScene \cite{lin2024instructscene}, SceneWeaver \cite{yang2025sceneweaver}, Ctrl-Room \cite{fang2025ctrlroom}, and ReSpace \cite{bucher2025respace}—significantly improve semantic controllability by allowing users to specify layout requirements through natural language instead of manually designing optimization objectives or spatial constraints. Despite these advances, existing methods are almost exclusively developed and evaluated on Manhattan-style environments, where room geometry can be approximated by axis-aligned boundaries.

\subsection{Challenges of Non-Manhattan Scene Generation}

Real-world indoor environments frequently deviate from the Manhattan-world assumption, exhibiting arbitrary wall orientations and irregular room boundaries. Such non-Manhattan geometries have been extensively studied in indoor layout estimation, floor-plan reconstruction, and scene understanding \cite{zou2018layoutnet,sun2019horizonnet,kalervo2019cubicasa5k,zheng2020structured3d}, leading to increasingly accurate geometric representations of complex indoor spaces. However, these studies primarily focus on recovering room structures rather than generating object layouts conditioned on user intent. Consequently, although recent LLM-based scene synthesis methods achieve impressive language understanding and semantic reasoning capabilities, they generally lack explicit modeling of non-orthogonal spatial relationships and polygonal room geometries. Applying them directly to irregular floor plans often results in physically implausible object placements and inconsistent orientations. However, our method learns features under non-Manhattan layouts by constructing spatial priors, enabling text-driven scene synthesis to generalize to general non-Manhattan environments.

\begin{figure*}
    \centering
    \includegraphics[width=0.90\linewidth]{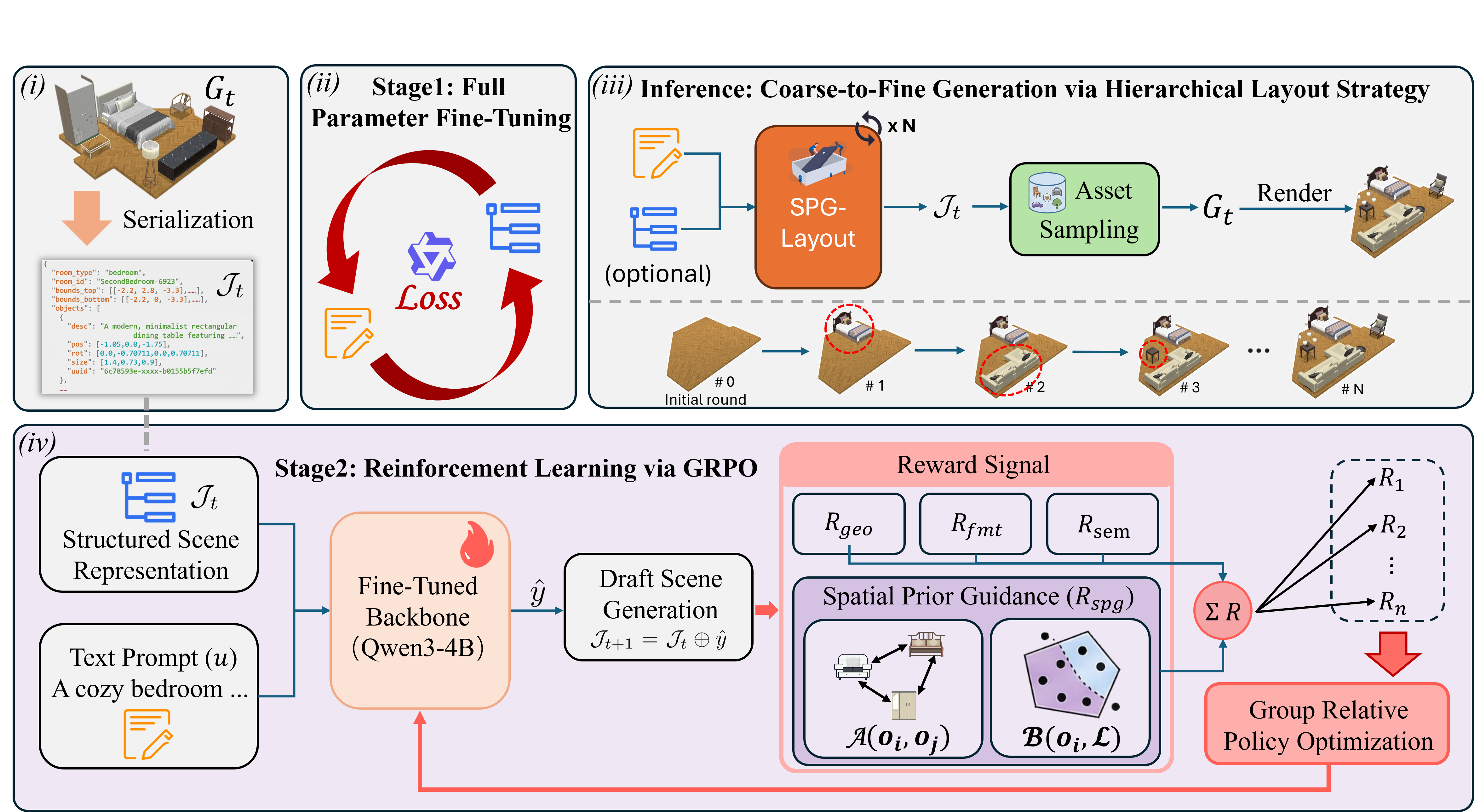}
    \caption{\textbf{Overview of the proposed SPG-Layout framework.} (\textit{\romannumeral 1}) Provides an intuitive comprehension of structured scene representation (Sec.~\ref{sec:3.1}). (\textit{\romannumeral 2}) Structural alignment by full parameter fine-tuning (Sec.~\ref{sec:train} stage 1). (\textit{\romannumeral 3}) Inference Pipeline illustrates the hierarchical layout strategy for progressive generation (Sec.~\ref{sec:hls}). (\textit{\romannumeral 4}) Reinforcement Learning optimizes the model using a multi-objective reward function via GRPO (Sec.~\ref{sec:train} stage 2).}
    
    \label{fig:pipeline}
    \vspace{-3mm}
\end{figure*}

\section{Preliminaries and Problem Statement} \label{sec:3}

\subsection{Preliminaries}\label{sec:3.1}
We represent a 3D indoor scene using a \textit{Structured Scene Representation (SSR)}, which disentangles the scene into an architectural layout and a set of object instances.

\noindent\textbf{Layout Definition.}
Let the room layout $\mathcal{L}$ be defined by the ceiling and floor boundaries, each represented as an ordered sequence of vertices,
$\mathcal{V}^{top} = (\mathbf{v}^{top}_1, \dots, \mathbf{v}^{top}_m)$ and
$\mathcal{V}^{bottom} = (\mathbf{v}^{bottom}_1, \dots, \mathbf{v}^{bottom}_m)$, where
$\mathbf{v}^{top}_i, \mathbf{v}^{bottom}_i \in \mathbb{R}^3$.
Adjacent vertices in each sequence define the polygon edges, and the first and last vertices are connected to form closed polygons in Euclidean space. These boundaries specify the room envelope and impose structural constraints for object placement.

\noindent\textbf{Object Parameterization.}
To support open-vocabulary synthesis, we represent each object $o_i$ as a tuple that decouples semantics from geometry:
\begin{equation}
    o_i = (d_i, \mathbf{s}_i, \mathbf{p}_i, \mathbf{q}_i),
\end{equation}
where $d_i$ denotes the natural language description (e.g., ``a leather sofa''), and $\mathbf{s}_i \in \mathbb{R}^3$, $\mathbf{p}_i \in \mathbb{R}^3$, $\mathbf{q}_i \in \mathbb{R}^4$ correspond to the 3D scale, position, and rotation, respectively. This factorization enables asset-agnostic synthesis via cross-modal retrieval.

\subsection{Problem Statement}\label{sec:3.2}
We formulate text-driven indoor scene synthesis as a conditional sequence generation task.
At generation step $t$, the scene state is represented as
$G_t = (\tau, \mathcal{L}, \mathcal{O}_t)$,
where $\tau$ denotes the semantic room type inferred from the user prompt $u$, $\mathcal{L}$ represents the geometric layout, and $\mathcal{O}_t = \{o_i\}_{i=1}^n$ is the set of instantiated objects.

To leverage the generative capabilities of LLMs, we introduce a serialization function
$\mathcal{S}: \mathcal{G} \rightarrow \mathcal{J}$,
which maps the structured scene state space $\mathcal{G}$ into a discrete SSR sequence space $\mathcal{J}$.
Given the user prompt $u$ and the serialized context $\mathcal{J}_t = \mathcal{S}(G_t)$ of the current state, our objective is to generate a layout-compliant structural increment $\hat{y} \in \mathcal{J}$ by maximizing the conditional likelihood:
\begin{equation}
    \hat{y} = \arg\max_{y \in \mathcal{J}} p_\theta(y \mid u, \mathcal{J}_t).
\end{equation}
The SSR is then updated in the serialized space as
$\mathcal{J}_{t+1} = \mathcal{J}_t \oplus \hat{y}$,
where $\oplus$ denotes the SSR update operator.
The updated structured scene state $G_{t+1}$ is subsequently sampled from $\mathcal{J}_{t+1}$, after which concrete assets are retrieved and instantiated to obtain the executable 3D scene.
The generation process is initialized at $t=0$ with
$G_0 = (\tau(u), \mathcal{L}, \emptyset)$,
whose serialized form is $\mathcal{J}_0 = \mathcal{S}(G_0)$.

\section{Framework} \label{sec:method}
This section details our 3D indoor scene synthesis framework (see Fig.~\ref{fig:pipeline}). We first describe the non-Manhattan data generation pipeline (Sec.~\ref{sec:4}). Then, we introduce the Spatial Prior Guidance mechanism (Sec.~\ref{sec:method1}) and the two-stage training process (Sec.~\ref{sec:train}). Finally, we present the hierarchical strategy used for inference in Sec.~\ref{sec:hls}.

\subsection{Non-Manhattan Data Collection} \label{sec:4}

\begin{figure*}
    \centering
    \includegraphics[width=0.95\linewidth]{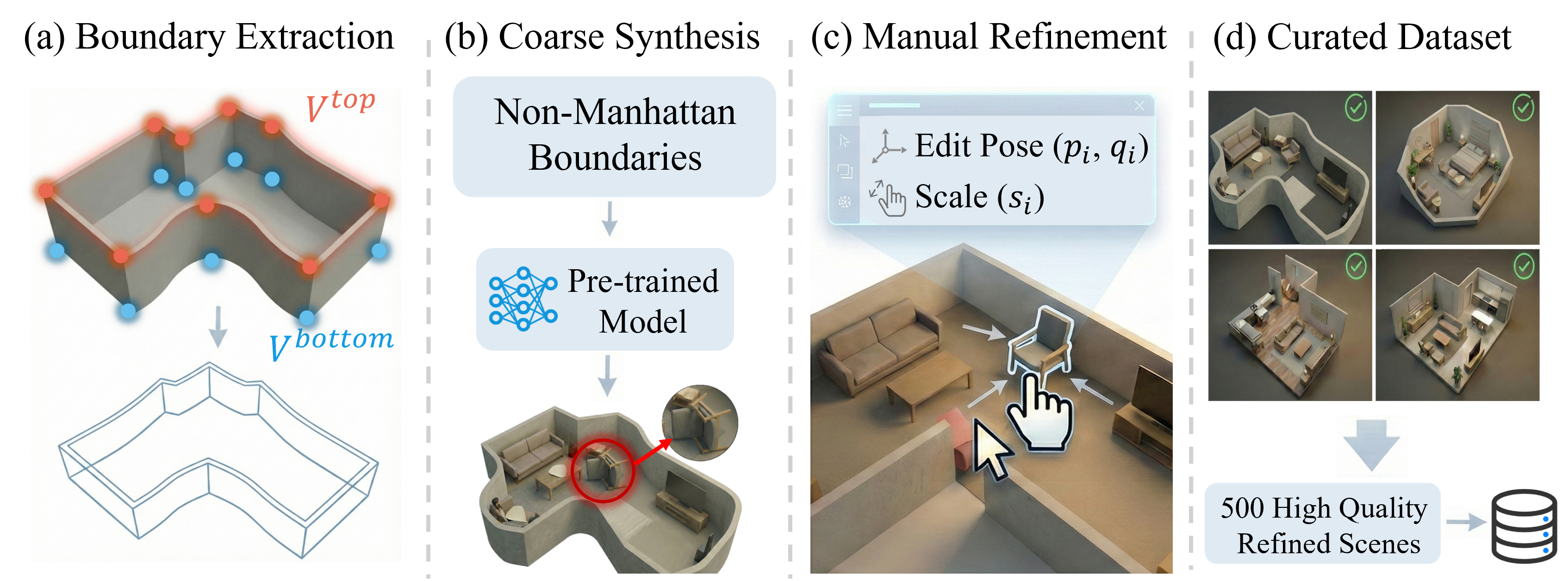}
    \caption{\textbf{Semi-Automated Data Generation Pipeline.} We extract geometric boundaries (a) and generate coarse scenes (b), followed by manual refinement of object attributes (c) to construct a curated dataset of 500 non-Manhattan scenes (d).}
    \label{fig:dataset}
    \vspace{-3mm}
\end{figure*}

To facilitate our research on non-Manhattan Environments, we present a semi-automated data generation pipeline, illustrated in Fig.~\ref{fig:dataset}, to construct a dataset of non-Manhattan 3D indoor scenes. 
Specifically, we first curate a large-scale collection of raw floor plan templates from web repositories and extract their geometric boundaries, formulated as $\mathcal{L}$. 
To populate these empty rooms, we utilize a model pre-trained on Manhattan datasets (specifically, the ReSpace~\cite{bucher2025respace} model integrated with our HLS strategy) to synthesize coarse non-Manhattan scenes. 
Since direct transfer may introduce artifacts, we subsequently employ a manual refinement process where annotators manually adjust the pose and scale of objects to ensure physical plausibility and semantic coherence. 
This pipeline ultimately yields a curated dataset of $500$ high-quality refined scenes. See more details in Sec.~\ref{supsec:0}.

\begin{figure}
    \centering
    \includegraphics[width=0.92\linewidth]{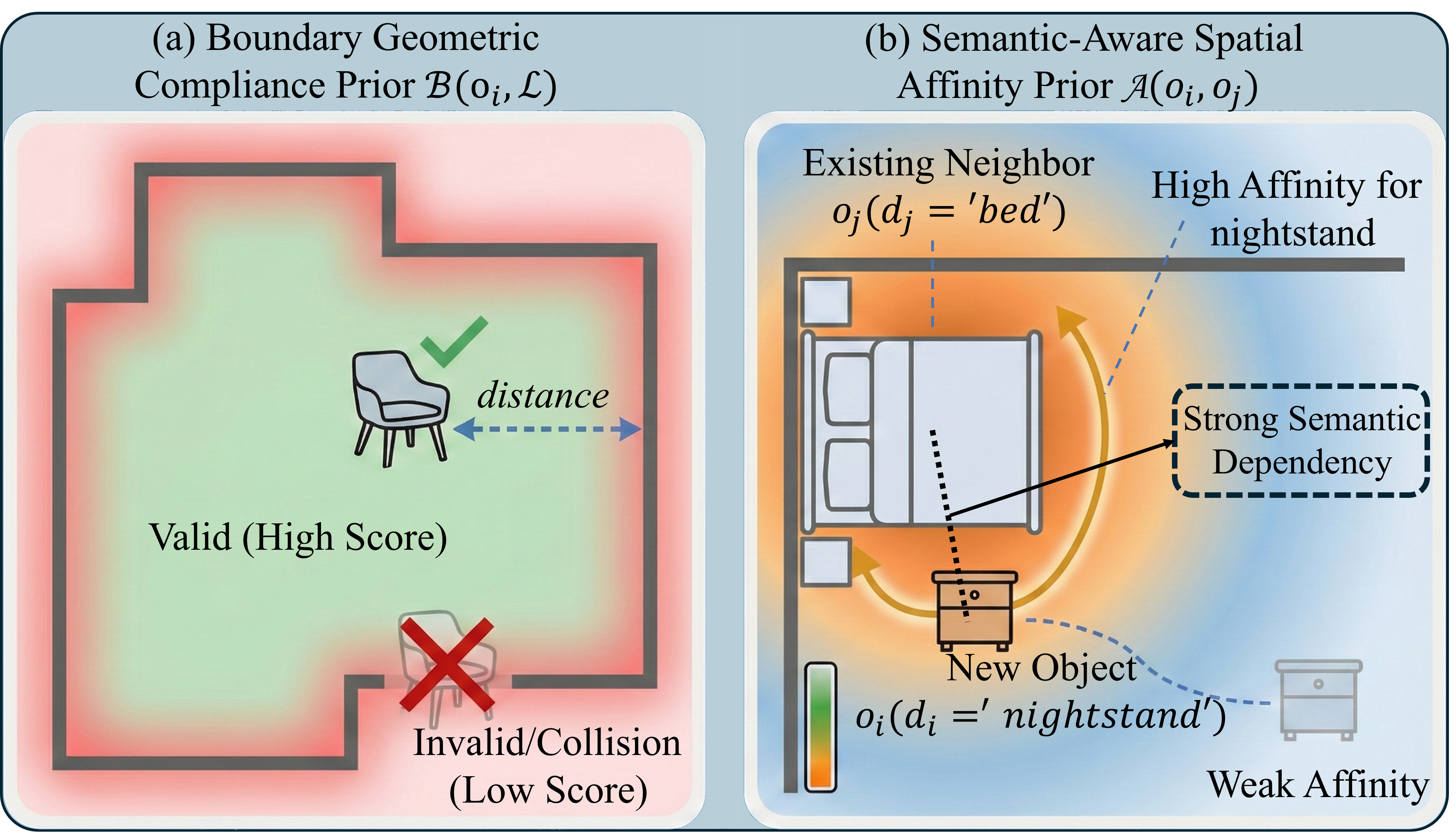}
    \caption{\textbf{Mechanism of Spatial Prior Guidance (SPG)}. SPG explicitly integrates geometric boundary priors and semantic co-occurrence affinities to provide computable reward signals for reinforcement learning-based scene generation.}
    \label{fig:spg}
    \vspace{-3mm}
\end{figure}

\subsection{Spatial Prior Guidance} \label{sec:method1}

While LLMs excel at semantic decomposition---parsing user prompts into object tuples $o_i = (d_i, \mathbf{s}_i, \mathbf{p}_i, \mathbf{q}_i)$---they exhibit significant limitations in spatial positioning $\mathbf{p}_i$ and geometric reasoning. Existing autoregressive approaches directly regress object parameters, effectively memorizing coordinate distributions from training data. In non-Manhattan environments with irregular boundaries $\mathcal{L}$, these memorized distributions fail to generalize: positions that are valid in Manhattan settings may fall outside the valid polygon, causing collisions or implausible placements. Moreover, without explicit spatial priors, the model lacks the inductive bias to capture conditional dependencies $P(\mathbf{o}_i \mid \mathbf{o}_j)$ between semantically affiliated objects (\textit{e.g.}, a nightstand adjacent to a bed) in irregular geometries.

To address these challenges, we introduce \textbf{\textit{Spatial Prior Guidance (SPG)}}. Instead of relying solely on the textual likelihood, we explicitly model the statistical priors of object-boundary relationships and object-object affinities, as illustrated in Fig.~\ref{fig:spg}.
Formally, we quantify the SPG as a scoring function $R_{\text{spg}}$ that evaluates the spatial validity of the generated position:
\begin{equation} \label{eq:spg_reward}
\begin{split}
    R_{\text{spg}}(\mathbf{o}_i, \mathcal{L}, \mathcal{O}_{<i}) = \alpha \cdot \mathcal{B}(\mathbf{o}_i, \mathcal{L}) & \\
    + \beta \cdot \sum_{o_j \in \mathcal{O}_{<i}} w_j \cdot \mathcal{A}(\mathbf{o}_i, \mathbf{o}_j) & ,
\end{split}
\end{equation}
\noindent where $\mathcal{B}(\cdot)$ measures the object-architecture prior consistency, and $\mathcal{A}(\cdot)$ calculates the semantic-aware spatial affinity between the current object and its neighbors in the previously generated set $\mathcal{O}_{<i}$. Sensitivity coefficients $\alpha$ and $\beta$ balance the contribution of each prior. The weight $w_j$ is computed via an attention mechanism:
\begin{equation} \label{eq:attn_weight}
    w_j = \frac{\exp(\mathcal{A}(\mathbf{o}_i, \mathbf{o}_j)/\tau)}{\sum_{o_k \in \mathcal{O}_{<i}} \exp(\mathcal{A}(\mathbf{o}_i, \mathbf{o}_k)/\tau)},
\end{equation}
where $\tau$ is a learnable temperature parameter. Unlike naive averaging, which dilutes the spatial signal with noise from weakly related objects, this attention-weighted aggregation dynamically assigns higher weights to semantically stronger spatial anchors (e.g., the bed as the primary anchor for a nightstand), while still retaining information from all previously placed objects. 
This formulation transforms abstract spatial constraints into a computable metric. $R_{\text{spg}}$ serves as an intrinsic reward signal that guides policy optimization in the reinforcement learning stage. We emphasize that $R_{\text{spg}}$ provides the model with directional guidance toward human-preferred spatial distributions rather than exposing ground-truth answers; combined with the KL divergence penalty in GRPO, this prevents reward hacking and ensures the model learns generalizable spatial reasoning rather than overfitting to specific evaluation criteria.

\subsection{Two-Stage Training Strategy} \label{sec:train}

We propose a two-stage training pipeline to adapt a general-purpose LLM for 3D scene layout generation. The first stage aligns the model to the SSR format via Supervised Fine-Tuning, and the second stage enforces geometric constraints via Reinforcement Learning with Group Relative Policy Optimization (GRPO). 
\noindent\textbf{Stage 1: Supervised Fine-Tuning for Structural Alignment.}

A key requirement for text-driven scene synthesis is adherence to the SSR format defined in Sec.~\ref{sec:3.1}. We fine-tune the Qwen3-4B backbone on SSR-3DFRONT~\cite{bucher2025respace}—a structured version of 3D-FRONT~\cite{fu20213d_front}. We use full-parameter fine-tuning rather than LoRA~\cite{hu2022lora}, as LoRA achieves only a 78\% format success rate compared to 99\% for full fine-tuning after training for 100 epochs. Such high format compliance is necessary for the RL stage to correctly parse and optimize object parameters.

\noindent\textbf{Stage 2: Reinforcement Learning via GRPO.}

While the first stage enables valid syntax generation, it does not enforce continuous physical constraints such as object intersections or boundary violations. We further optimize the SFT model using GRPO to improve geometric plausibility. 

To balance geometric consistency, format compliance, semantic matching, and spatial reasoning, we employ a multi-objective reward function, as defined in Eq.~\ref{eq:rewardF}:
\begin{equation}
\label{eq:rewardF}
\begin{aligned}
R_{total} &=
\begin{cases}
R_{fmt},
& R_{fmt} \neq 1, \\[4pt]

\gamma_1 R_{geo}
+ \gamma_2 R_{sem} \\[2pt]
\qquad
+ \dfrac{\gamma_3}{N}
\sum_{i=1}^{N}
R_{spg},
& \text{otherwise}.
\end{cases}
\end{aligned}
\end{equation}
\noindent where $\gamma_i$ ($i \in \{1,2,3\}$) represents the weight coefficient for each reward component, and $N$ denotes the number of objects in the synthesized scene. The specific definitions of each reward component are as follows:

\noindent\textbf{Geometric Consistency Reward ($R_{geo}$).}
We design the geometric reward to enforce physical constraints by minimizing an energy-based cost. The \textit{geometric energy} $\mathcal{E}_{geo}$ measures the severity of boundary violations and object collisions:
\begin{equation}
\mathcal{E}_{geo} = \sum_{i} \underbrace{\| \mathbf{V}_i \odot (\mathbf{1} - \mathbf{V}_{env}) \|_1}_{\text{Boundary Violation}} + \sum_{i < j} \underbrace{\| \mathbf{V}_i \odot \mathbf{V}_j \|_1}_{\text{Collision}},
\end{equation}
\noindent where $\mathbf{V}_i$ denotes the voxelized grid of the $i$-th object, and $\mathbf{V}_{env}$ represents the environment layout (with $1$ indicating valid interior space). $\odot$ denotes the Hadamard product, and $\| \cdot \|_1$ is the $L_1$ norm measuring the volume of overlap.
The final reward is obtained by normalizing this energy via exponential normalization: $R_{geo} = \exp ( - \mathcal{E}_{geo} / \sigma_{geo} )$. This mapping ensures that collision-free layouts receive a reward approaching $1$, while invalid configurations yield smooth penalties to facilitate gradient-based optimization.

\noindent\textbf{Format Compliance Reward ($R_{fmt}$).}
To guarantee that the generated outputs constitute a valid Structured Scene Representation (SSR) as defined in Sec.~\ref{sec:3.1}, we impose a strict structural constraint.
Let $\Phi(o_i)$ be a validation function that returns $1$ if and only if the generated object $o_i$ can be successfully parsed into the tuple $(d_i, \mathbf{s}_i, \mathbf{p}_i, \mathbf{q}_i)$ with correct dimensionality ($\mathbf{s}_i, \mathbf{p}_i \in \mathbb{R}^3, \mathbf{q}_i \in \mathbb{R}^4$).
The reward is the average validity over $N$ generated objects: 
\begin{equation}
R_{fmt} = \frac{1}{N} \sum_{i=1}^{N} \Phi(o_i).
\end{equation}
This ensures the model strictly adheres to the defined
schema, preventing malformed outputs that would fail the
subsequent rendering process.

\noindent\textbf{Semantic Matching Reward ($R_{sem}$).}
We quantify semantic fidelity by measuring the recall of mandatory attribute keywords, complementing latent-space metrics. Let $\mathcal{K}$ denote the set of keywords extracted from the user prompt (e.g., $\{\text{``red'', ``leather''}\}$), and $\mathcal{W}_i$ be the set of tokens in the generated object description. The reward is defined as the average keyword coverage:
\begin{equation}
r_{i} = \frac{1}{|\mathcal{K}|} \sum_{w \in \mathcal{K}} \mathbb{I}[w \in \bigcup_i \mathcal{W}_i],\quad R_{sem}=\frac{1}{N} \sum_{i} r_{i}.
\end{equation}
This metric explicitly penalizes the omission of fine-grained constraints, ensuring the synthesized scene textually aligns with specific user instructions. Note that, if $|\mathcal{K}|=0$, this term is excluded from the overall reward.

\subsection{Hierarchical Layout Strategy} \label{sec:hls}
While SPG-Layout excels in incremental object placement within existing scenes, text-driven 3D scene synthesis necessitates the ability to generate holistic scenes from scratch. In the absence of established spatial references, the unprioritized placement of smaller objects risks fragmenting the continuous free space. This fragmentation creates severe geometric constraints, often rendering the subsequent accommodation of larger structures infeasible due to collision conflicts. Conversely, prioritizing larger entities establishes a spatial backbone, ensuring sufficient allocation for dominant structures while allowing smaller items to be flexibly positioned within the remaining interstices.

\begin{figure}
    \centering
    \includegraphics[width=0.99\linewidth]{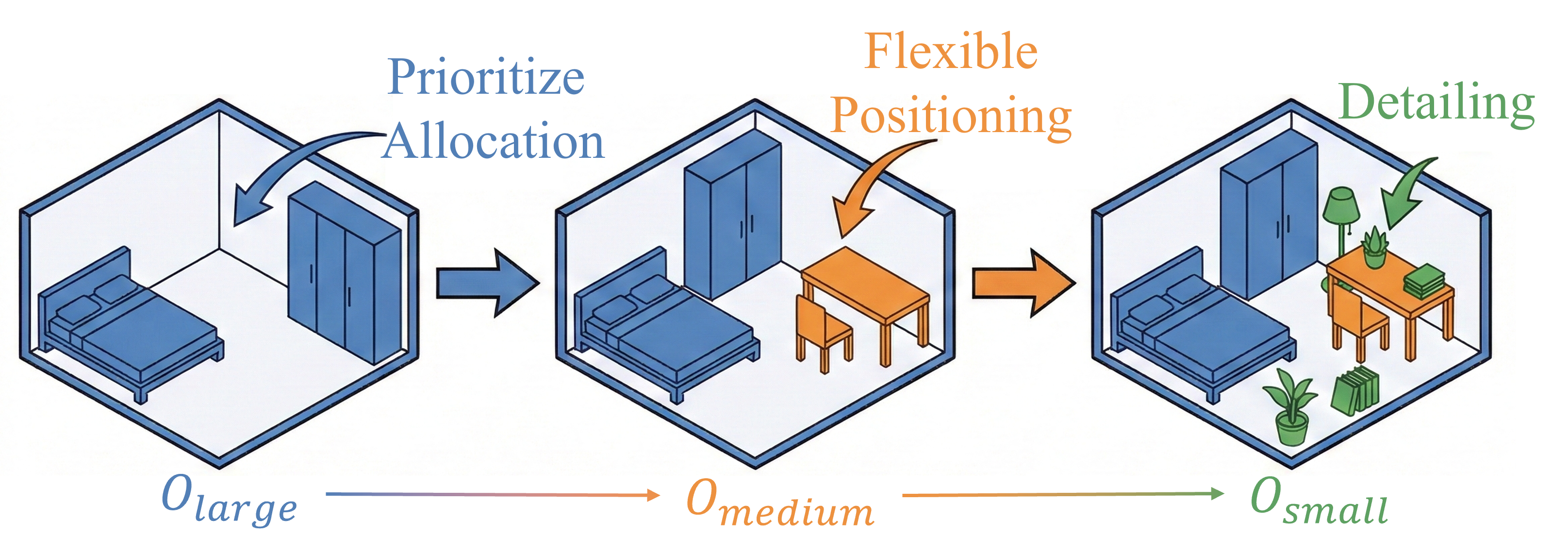}
    \caption{\textbf{The Illustration of HLS.} }
    \label{fig:hls}
    \vspace{-5mm}
\end{figure}

Drawing inspiration from human cognitive strategies in interior design—where dominant furniture pieces (e.g., beds, sofas) are positioned prior to supplementary items (e.g., stools, side tables)—we introduce the Hierarchical Layout Strategy (HLS). Specifically, we categorize objects into three distinct tiers based on area occupancy: large ($\mathcal{O}_{\text{large}}$, $area > 1.2$), medium ($\mathcal{O}_{\text{medium}}$, $0.3 \le area \le 1.2$) and small ($\mathcal{O}_{\text{small}}$, $area < 0.3$). HLS strictly enforces a descending order of placement ($\mathcal{O}_{\text{large}} \rightarrow \mathcal{O}_{\text{medium}} \rightarrow \mathcal{O}_{\text{small}}$), prioritizing the localization of large objects before determining the coordinates for smaller entities (As shown in Fig.~\ref{fig:hls}). 

Furthermore, HLS is adapted as a dynamic conflict resolution mechanism during single-object addition. When the initial scene layout is suboptimal and leads to collisions between the SPG-Layout prediction and existing objects, HLS temporarily retracts all existing objects smaller than the new entity into a candidate pool. The system then re-optimizes the layout by placing all objects in the pool (including the new one) following the hierarchical size-descending protocol. This approach facilitates the seamless integration of new objects while maximally preserving the structural integrity and semantic consistency of the original scene.

\section{Experiments}\label{sec:experiments}
We evaluate SPG-Layout on two tasks in non-Manhattan environments: 1) \textbf{Single Object Addition:} inserting new objects into existing layouts via text instructions; 2) \textbf{Full Scene Synthesis:} generating complete layouts from text descriptions alone. We also benchmark against methods restricted to orthogonal geometries in standard Manhattan settings. Finally, we conduct ablation studies to validate each proposed module.

\begin{table*}[tbh]
\centering
\caption{\textbf{Quantitative comparison on layout generation.} We report results as \textit{\textbf{Single Object Addition}} / \textit{\textbf{Full Scene Synthesis}} tasks. Layout Violations metrics are multiplied by $10^3$ for readability. $\uparrow$ indicates higher is better, and $\downarrow$ indicates lower is better. \textbf{Bold} indicates the best performance for each setting respectively. ``n/a'' denotes inapplicable metrics.  
}
\label{tab:1}
\renewcommand{\arraystretch}{1.25}
\setlength{\tabcolsep}{3pt}

\resizebox{\textwidth}{!}{%
\begin{tabular}{l l c c c c c c c}
\toprule
\multirow{2}{*}{\textbf{Category}} & \multirow{2}{*}{\textbf{Method}} & \multicolumn{3}{c}{\textbf{Layout Violations}} & \multicolumn{4}{c}{\textbf{Layout Fidelity}} \\
\cmidrule(lr){3-5} \cmidrule(lr){6-9}
 & & {$\downarrow$ \textbf{OOB}$_{\times 10^3}$} & {$\downarrow$ \textbf{MBL}$_{\times 10^3}$} & {$\downarrow$ \textbf{VBL}$_{\times 10^3}$} & {$\uparrow$ \textbf{OOR}(\%)} & {$\uparrow$ \textbf{OAR}(\%)} & {$\uparrow$ \textbf{ATR}(\%)} & {$\uparrow$ \textbf{PMS}(\%)} \\
\midrule

\multirow{7}{*}{\textbf{Bed.}} 
  & LayoutGPT ~\cite{feng2023layoutgpt}
    & 850.91 / 2243.30 & 251.20 / 662.26 & 1102.11 / 2905.56 & 8.15 / 6.02 & 11.39 / 8.42 & 16.03 / 11.85 & n/a / n/a \\
  & LayoutVLM ~\cite{sun2024layoutvlm}
    & 192.22 / 506.76 & 193.58 / 510.36 & 385.80 / 1017.12 & 34.64 / 25.61 & 40.14 / 29.67 & 47.91 / 35.41 & n/a / n/a \\
  & InstructScene ~\cite{lin2024instructscene}
    & 338.22 / 891.68 & 414.73 / 1093.38 & 752.95 / 1985.06 & 17.16 / 12.69 & 25.36 / 18.74 & 32.55 / 24.06 & n/a / n/a \\

  & ATISS~\cite{paschalidou2021atiss} 
    & 237.7 / 523.2 & 156.7 / 330.8 & 394.4 / 853.2 & 33.13 / 23.87 & 41.54 / 29.78 & 56.78 / 34.90 & 52.00 / n/a \\
  & MiDiffusion~\cite{hu2024mixed}        
    & 158.0 / 422.3 & 114.1 / 298.8 & 272.2 / 721.1 & 36.87 / 28.78 & 47.89 / 35.94 & 61.52 / 39.02 & 50.00 / n/a \\
  & ReSpace~\cite{bucher2025respace}  
    & 53.22 / 261.5 & 11.17 / 167.2 & 64.39 / 428.7 & 58.32 / 60.69 & 74.01 / 71.81 & \textbf{89.68} / 77.38 & \textbf{91.00} / 87.00 \\
  \rowcolor{gray!20} & \textbf{SPG-Layout (Ours)}        
    & \textbf{1.56} / \textbf{29.30} & \textbf{1.63} / \textbf{30.39} & \textbf{3.19} / \textbf{59.69} & \textbf{63.64} / \textbf{62.81} & \textbf{79.42} / \textbf{75.71} & 88.71 / \textbf{81.57} & \textbf{91.00} / \textbf{89.00} \\
\midrule

\multirow{7}{*}{\textbf{Liv.}} 
  & LayoutGPT ~\cite{feng2023layoutgpt}
    & 724.85 / 1910.96 & 213.99 / 564.15 & 938.83 / 2475.10 & 8.48 / 6.27 & 11.86 / 8.77 & 16.68 / 12.33 & n/a / n/a \\
  & LayoutVLM ~\cite{sun2024layoutvlm}
    & 163.74 / 431.68 & 164.91 / 434.75 & 328.65 / 866.43 & 36.06 / 26.65 & 41.78 / 30.88 & 49.86 / 36.86 & n/a / n/a \\
  & InstructScene ~\cite{lin2024instructscene}
    & 288.12 / 759.58 & 353.29 / 931.40 & 641.41 / 1690.98 & 17.86 / 13.20 & 26.39 / 19.51 & 33.88 / 25.04 & n/a / n/a \\

  & ATISS~\cite{paschalidou2021atiss} 
    & 197.0 / 504.3 & 131.2 / 254.8 & 328.2 / 759.1 & 36.13 / 29.56 & 43.54 / 31.59 & 58.78 / 36.88 & 56.00 / n/a \\
  & MiDiffusion~\cite{hu2024mixed}        
    & 142.3 / 352.6 & 99.76 / 237.2 & 242.0 / 589.8 & 41.08 / 31.73 & 48.70 / 32.21 & 67.52 / 37.94 & 57.00 / n/a \\
  & ReSpace~\cite{bucher2025respace}  
    & 31.01 / 190.5 & 13.29 / 65.20 & 44.30 / 255.7 & 61.21 / 63.43 & 65.99 / 65.38 & 89.05 / 83.41 & \textbf{90.00} / \textbf{89.00} \\
  \rowcolor{gray!20} & \textbf{SPG-Layout (Ours)}        
    & \textbf{1.47} / \textbf{4.05} & \textbf{1.30} / \textbf{6.06} & \textbf{2.77} / \textbf{10.11} & \textbf{66.69} / \textbf{65.54} & \textbf{67.63} / \textbf{69.66} & \textbf{89.60} / \textbf{84.37} & \textbf{90.00} / \textbf{89.00} \\
\midrule

\multirow{8}{*}{\textbf{All}}
  & LayoutGPT ~\cite{feng2023layoutgpt}
    & 787.88 / 2077.12 & 232.59 / 613.20 & 1020.47 / 2690.33 & 8.31 / 6.15 & 11.63 / 8.59 & 16.35 / 12.09 & n/a / n/a \\
  & LayoutVLM ~\cite{sun2024layoutvlm}
    & 177.98 / 469.22 & 179.25 / 472.55 & 357.23 / 941.77 & 35.35 / 26.13 & 40.96 / 30.28 & 48.89 / 36.13 & n/a / n/a \\
  & InstructScene ~\cite{lin2024instructscene}
    & 313.17 / 825.63 & 384.01 / 1012.39 & 697.18 / 1838.02 & 17.51 / 12.95 & 25.87 / 19.12 & 33.21 / 24.55 & n/a / n/a \\

  & ATISS~\cite{paschalidou2021atiss}
    & 221.7 / 631.4 & 114.5 / 108.5 & 336.1 / 739.8 & 35.92 / 27.56 & 44.04 / 23.66 & 57.09 / 28.76 & 55.00 / n/a \\
  & MiDiffusion~\cite{hu2024mixed}
    & 153.5 / 327.4 & 102.9 / 87.10 & 256.4 / 414.5 & 39.77 / 29.02 & 47.00 / 26.88 & 65.10 / 31.23 & 53.00 / n/a \\
  
  & ReSpace~\cite{bucher2025respace}
    & 34.33 / 218.4 & 12.26 / 91.94 & 46.60 / 310.3 & 60.24 / 60.61 & 73.17 / 69.07 & \textbf{89.23} / 82.51 & \textbf{90.00} / 90.00 \\
  \rowcolor{gray!20} & \textbf{SPG-Layout (Ours)}
    & \textbf{0.59} / \textbf{14.93} & \textbf{1.12} / \textbf{14.36} & \textbf{1.71} / \textbf{29.29} & \textbf{66.18} / \textbf{65.71} & \textbf{76.68} / \textbf{74.78} & 89.09 / \textbf{83.63} & 89.00 / \textbf{92.00} \\
\bottomrule
\end{tabular}%
}
\vspace{-3mm}
\end{table*}

\subsection{Experimental Setup}

We evaluate on the 500 non-Manhattan scenes from Sec.~\ref{sec:4}, partitioned into `bed’ (252), `liv’ (148), and `all’ (500) splits following ReSpace~\cite{bucher2025respace}. For Manhattan comparison, we sample an equivalent subset from SSR-3DFRONT~\cite{bucher2025respace}. We report Layout Violation metrics (OOB, MBL, VBL) and Layout Fidelity metrics (OOR, OAR, ATR, PMS), with definitions in Sec.~\ref{supsec:1}. Independent SceneEval~\cite{tam2026sceneeval} results are in Sec.~\ref{supsec:5}.

\subsection{Main Results}
The model is trained on SSR-3DFRONT~\cite{bucher2025respace} for 110 epochs, comprising 100 epochs of supervised fine-tuning and 10 epochs of reinforcement learning. GRPO is adopted with a group size of 8, where $\alpha=0.55$ and $\beta=0.45$ in Eq.~\ref{eq:spg_reward}, and $\gamma_{1\text{-}3}=0.3, 0.3, 0.4$ in Eq.~\ref{eq:rewardF}.

\noindent\textbf{Single Object Addition.}
Tab.~\ref{tab:1} shows SPG-Layout outperforms baselines by a significant margin. In the `bedroom' subset, OOB, MBL, and VBL are reduced by 51.66, 9.54, and 61.2, respectively. OOR and OAR improve by an average of 5.58 and 3.52 across all categories.

\begin{figure}
    \centering
    \includegraphics[width=0.90\linewidth]{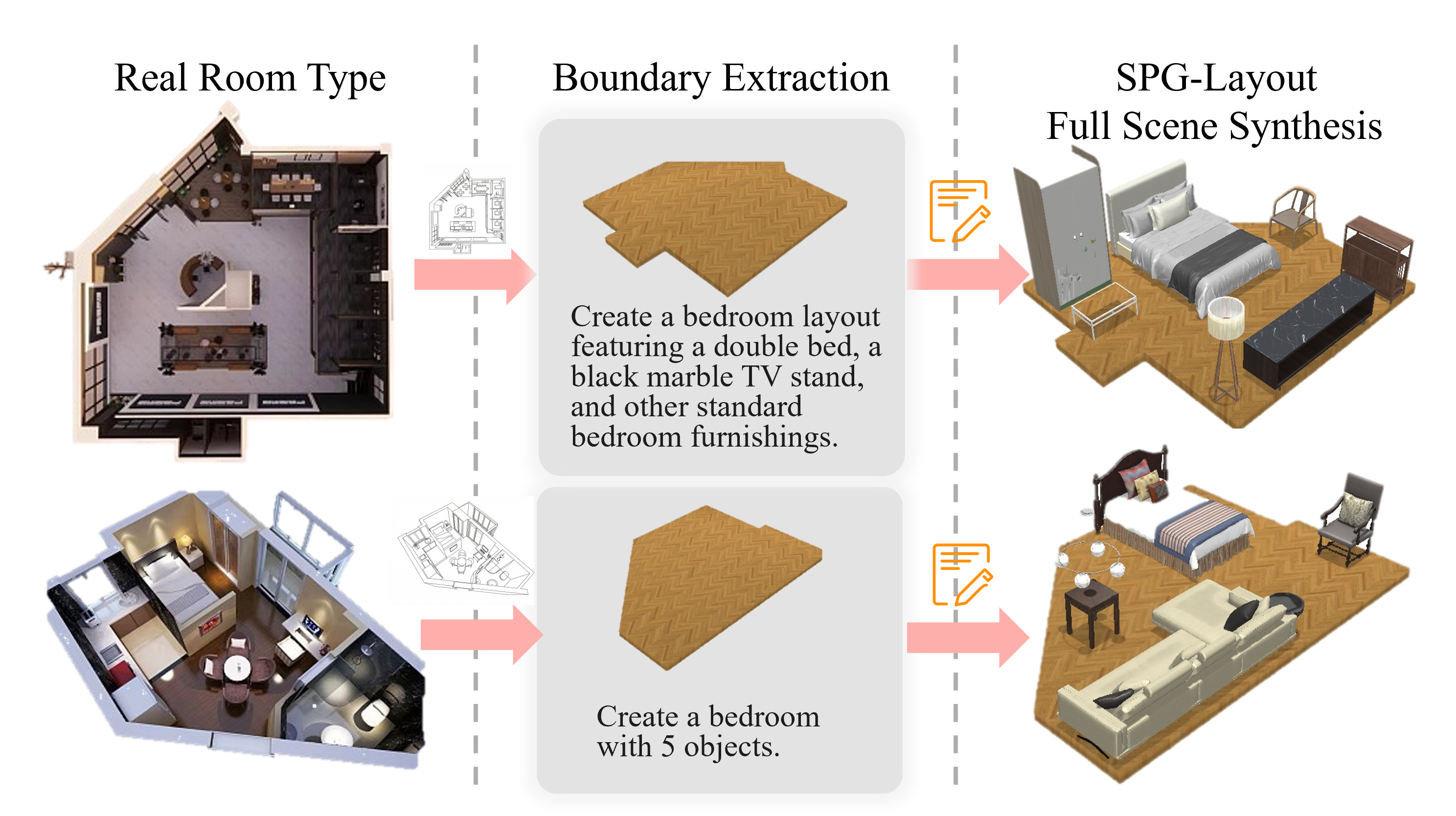}
    \caption{\textbf{Qualitative Results on non-Manhattan Dataset.} Given real-world room type, SPG-Layout generates complete scene layouts guided by specific user instructions.}
    \label{fig:full_exp}
    \vspace{-3mm}
\end{figure}

\begin{figure*}
    \centering
    \includegraphics[width=0.9\linewidth]{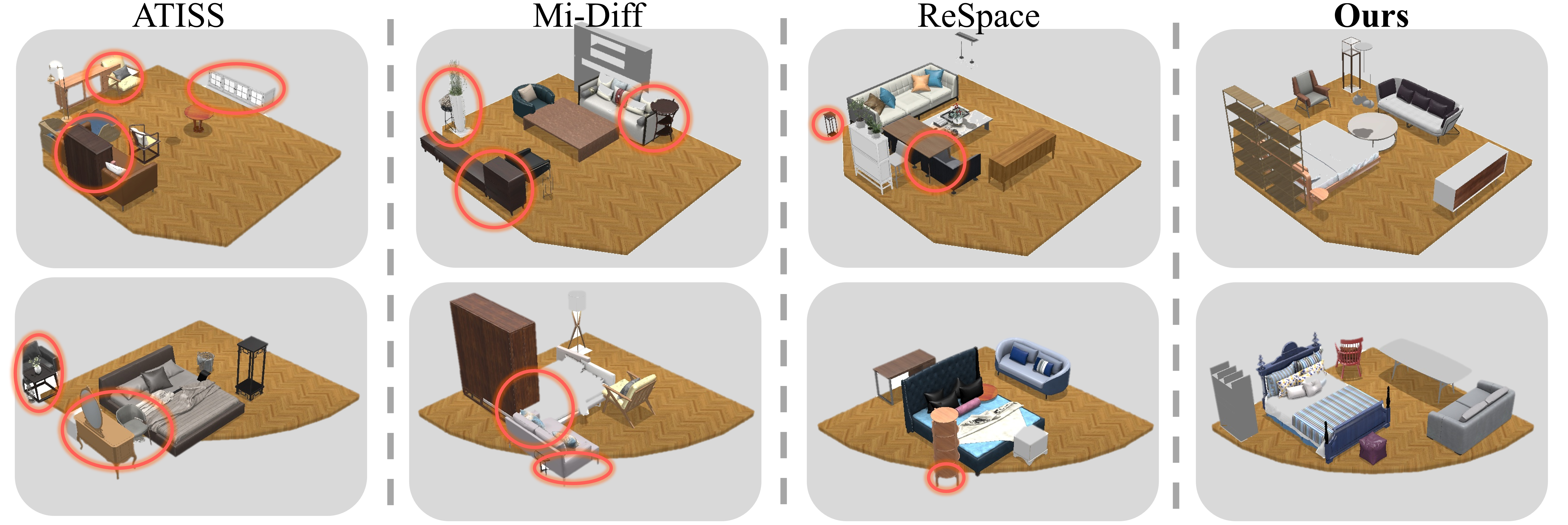}
    \caption{Qualitative results of layout generation in complex non-Manhattan environments compared to the existing methods. \textcolor{red!80}{Red circles} highlight instances of layout violations, such as object collisions or boundary penetrations.}
    \label{fig:vis}
    \vspace{-3mm}
\end{figure*}

\noindent\textbf{Full Scene Synthesis.}
Starting from an empty scene, the model must synthesize the entire layout from text alone. Tab.~\ref{tab:1} demonstrates layout violation reductions of 86\%, 96\%, and 91\% in the Bedroom, Living room, and All categories over the runner-up, while achieving SOTA fidelity. The attention-weighted SPG (Eq.~\ref{eq:spg_reward}) improves OOR from 63.39\% to 65.71\% in the All category compared to the averaging baseline, confirming that dynamically weighting spatial anchors yields more effective guidance. 
Qualitative results are shown in Fig.~\ref{fig:vis} and Fig.~\ref{fig:full_exp}.

\begin{table*}[t]
\centering
\caption{\textbf{Generalization results in conventional Manhattan environments.} $\uparrow$ indicates higher is better, and $\downarrow$ indicates lower is better. Layout Violations are multiplied by $10^3$ for readability. \textbf{Bold}  indicate the best performance.}
\label{tab:gen}
\renewcommand{\arraystretch}{1.2} 
\setlength{\tabcolsep}{4pt}       

\resizebox{\textwidth}{!}{%
\begin{tabular}{l *{3}{S[table-format=3.2]} *{4}{c}}
\toprule
\multirow{2}{*}{\textbf{Method}} & \multicolumn{3}{c}{\textbf{Layout Violations}} & \multicolumn{4}{c}{\textbf{Layout Fidelity}} \\
\cmidrule(lr){2-4} \cmidrule(lr){5-8}
& {$\downarrow$ \textbf{OOB}$_{\times 1\text{e}3}$} & {$\downarrow$ \textbf{MBL}$_{\times 1\text{e}3}$} & {$\downarrow$ \textbf{VBL}$_{\times 1\text{e}3}$} & {$\uparrow$ \textbf{OOR}(\%)} & {$\uparrow$ \textbf{OAR}(\%)} & {$\uparrow$ \textbf{ATR}(\%)} & {$\uparrow$ \textbf{PMS}(\%)} \\
\midrule

LayoutGPT ~\cite{feng2023layoutgpt}      & 1199.7  & 84.21  & 1284.0  & 9.87 & 12.78 & 15.32 & {n/a} \\
LayoutVLM ~\cite{sun2024layoutvlm}    & 78.6  & 84.34  & 162.9  & 33.78 & 39.24 & 46.45 & {n/a} \\
InstructScene ~\cite{lin2024instructscene}    & 211.45 & 125.71 & 337.16 & 23.38 & 33.79 & 38.20 & {n/a} \\
ATISS ~\cite{paschalidou2021atiss}      & 412.75  & 250.8  & 663.55  & 26.87 & 32.78 & 37.85 & {n/a} \\
Mi-Diff ~\cite{hu2024mixed}    & 422.3  & 298.8  & 721.1  & 35.06 & 41.37 & 44.98 & {n/a} \\
ReSpace ~\cite{bucher2025respace}    & 88.8 & 124.60 & 213.4 & 69.61 & 77.29 & 78.32 & 72.00 \\
  \rowcolor{gray!20}  \textbf{SPG-Layout (Ours)}                   & \textbf{8.12}  & \textbf{9.73}  & \textbf{17.85}  & \textbf{73.21} & \textbf{81.23} & \textbf{85.07} & \textbf{89.00} \\

\bottomrule
\end{tabular}%
}
\vspace{-2mm}
\end{table*}

\subsection{Generalization Capacity and User Study}
\noindent\textbf{Generalization Capacity.} To evaluate generalization, we test in standard Manhattan environments (Tab.~\ref{tab:gen}). Results show that our method outperforms existing approaches in both non-Manhattan and conventional Manhattan settings.

\noindent\textbf{User Study.} We conducted a user study to evaluate perceptual quality in non-Manhattan scenes. We generated 100 comparison sets, each containing our method's output alongside all baselines, presented in randomized and anonymized order. Twenty participants voted for the layout that best preserved spatial fidelity, object arrangement rationality, and consistency with the input description (Tab.~\ref{tab:user_study}). 

\begin{table}[h]
\centering
\caption{\textbf{User Study Results.}
Twenty participants blindly voted for the best layout among anonymized results in each comparison set.}
\label{tab:user_study}
\renewcommand{\arraystretch}{1.15}
\setlength{\tabcolsep}{5pt}

\resizebox{\linewidth}{!}{%
\begin{tabular}{lccccccc}
\toprule
\textbf{Method} 
& \textbf{LayoutGPT} 
& \textbf{LayoutVLM} 
& \textbf{InstructScene} 
& \textbf{ATISS} 
& \textbf{Mi-Diff} 
& \textbf{ReSpace} 
& \textbf{Ours} \\
\midrule
\textbf{Votes} 
& 34 
& 71 
& 48 
& 56 
& 83 
& 346 
& \textbf{1362} \\
\bottomrule
\end{tabular}%
}
\vspace{-5mm}
\end{table}

\subsection{Ablation Study}
We report two aggregated metrics: \textbf{avg\_LV} (mean of OOB, MBL, VBL) and \textbf{avg\_LF} (mean of OOR, OAR, ATR, PMS).

\noindent\textbf{Impact of Reward Signals.}
As shown in the second row of Tab.~\ref{tab:ablation}, the SPG module improves layout quality and reduces violations. Without SPG, \textbf{avg\_LF} of the three categories drops by 15.6\%, 12.8\% and 11.4\%, and the average \textbf{avg\_LV} increases by 50.74. The spatial prior of SPG enhances scene understanding and yields more reasonable layout coordinate predictions.
Ablation on geometric reward $R_{geo}$ (third row) reveals that removing $R_{geo}$ slightly raises \textbf{avg\_LV}. Its impact is negligible compared with HLS, indicating $R_{geo}$ is not a core factor for mitigating layout violations.

\noindent\textbf{Impact of Hierarchical Layout Strategy.} 
As shown in the fourth row of Tab.~\ref{tab:ablation}, HLS is critical for mitigating layout violations. The full model reduces \textbf{avg\_LV} by 100.73, 108.89, and 113.16 on the `bed', `liv', and `all' splits compared with the model without HLS. This indicates that the hierarchical constraints of HLS prevent unreasonable placement of small objects and effectively avoid collisions between objects.

\begin{table}[t]
\centering
\caption{\textbf{Ablation Study.} }
\label{tab:ablation}
\renewcommand{\arraystretch}{1.3} 
\setlength{\tabcolsep}{5pt}       

\resizebox{\linewidth}{!}{%
\begin{tabular}{l ccc ccc}
\toprule
\multirow{2}{*}{\textbf{Settings}} & \multicolumn{3}{c}{\textbf{avg\_LV} ($\downarrow$)} & \multicolumn{3}{c}{\textbf{avg\_LF} ($\uparrow$)} \\
\cmidrule(lr){2-4} \cmidrule(lr){5-7}

& \textbf{bed} & \textbf{liv} & \textbf{all} & \textbf{bed} & \textbf{liv} & \textbf{all} \\
\midrule

\textbf{baseline}  & 285.78 & 170.47 & 206.86 & 74.22 & 75.31 & 75.54 \\
SPG-Layout (\textbf{w/o} $R_{spg}$) & 88.09 & 54.77 & 75.41 & 65.18 & 67.24 & 69.45 \\
SPG-Layout (\textbf{w/o} $R_{geo}$) & 102.77 & 86.09 & 78.76 & 76.64 & 75.08 & 77.89 \\
SPG-Layout (\textbf{w/o} HLS) & 140.52 & 115.63 & 132.68 & 73.25 & 74.81 & 76.39 \\
 \rowcolor{gray!20} \textbf{SPG-Layout (full)} & \textbf{39.79} & \textbf{6.74} & \textbf{19.52} & \textbf{77.27} & \textbf{77.14} & \textbf{78.36} \\

\bottomrule
\end{tabular}%
}
\vspace{-3mm}
\end{table}

\section{Conclusion}
In this paper, we present \textbf{SPG-Layout}, a novel text-driven framework for 3D indoor scene synthesis that effectively addresses the limitations of existing methods in handling non-Manhattan environments. By integrating \textit{Spatial Prior Guidance} and \textit{Hierarchical Layout Strategy} within a two-stage training framework, our approach achieves unprecedented advancements on complex non-Manhattan benchmark. Furthermore, comprehensive experiments reveal that SPG-Layout not only leads state-of-the-art baselines in general but also far exceeds the performance of existing methods on standard Manhattan Environments.

\clearpage

\section*{Appendix}
\label{sec:appendix}

This technical appendix provides additional details of metrics, implementation details of our method, as well as experimental results that are omitted from the main body of this paper due to the page limit.

\section{Preliminary: Group Relative Policy Optimization (GRPO)}

GRPO is a reinforcement learning algorithm for post-training large language models (LLMs). Compared with Proximal Policy Optimization (PPO), GRPO removes the need for an additional value (critic) network by estimating advantages through relative reward comparison within a group of sampled responses. This design significantly reduces computational and memory overhead while maintaining stable policy optimization.

Given an input prompt $x$, the old policy $\pi_{\theta_{\mathrm{old}}}$ samples a group of $G$ responses,
\[
\{y_1,y_2,\ldots,y_G\}\sim\pi_{\theta_{\mathrm{old}}}(\cdot|x).
\]
Each response is assigned a scalar reward $r_i$, and its advantage is computed by group-wise normalization:
\[
A_i=
\frac{r_i-\operatorname{mean}(\{r_j\}_{j=1}^{G})}
{\operatorname{std}(\{r_j\}_{j=1}^{G})+\epsilon},
\]
where $\epsilon$ is a small constant for numerical stability.

GRPO optimizes the policy using the following clipped surrogate objective:
\begin{equation}
\label{eq:grpo}
\begin{aligned}
&\mathcal{L}_{\mathrm{GRPO}}(\theta) = \\
&\quad\mathbb{E}\Bigg[
\frac{1}{G}
\sum_{i=1}^{G}
\Bigg(
\min\Big(
r_i(\theta)A_i,\,
\mathrm{clip}\big(
r_i(\theta),
1-\varepsilon,
1+\varepsilon
\big)A_i
\Big)
\\
&\quad\quad-\beta
D_{\mathrm{KL}}
\big(
\pi_{\theta}
\parallel
\pi_{\mathrm{ref}}
\big)
\Bigg)
\Bigg],
\end{aligned}
\end{equation}
where
\[
r_i(\theta)
=
\frac{\pi_{\theta}(y_i|x)}
{\pi_{\theta_{\mathrm{old}}}(y_i|x)}
\]
denotes the importance sampling ratio, $\pi_{\mathrm{ref}}$ is the reference policy, $\varepsilon$ is the clipping coefficient, and $\beta$ controls the KL regularization strength. The clipped objective constrains the policy update, while the KL term prevents the optimized policy from deviating excessively from the reference model.

The above formulation is adopted as the reinforcement learning backbone in our method. Readers are referred to the original GRPO paper for a comprehensive derivation and theoretical analysis.

\section{Details of Non-Manhattan Data Collection}\label{supsec:0}
Our dataset is collected in a copyright-respecting manner. All floor plans and related materials are obtained from publicly accessible sources under appropriate usage conditions, and we only use the information necessary for research purposes. We do not redistribute any proprietary raw content, and all collected data are processed into derived representations (e.g., geometric layouts and annotations) that do not retain identifiable copyrighted elements.

Our prior extraction strategy is highly robust. Its main purpose is to prevent the model from simply memorizing absolute coordinates, and instead encourage it to learn the underlying distributions of object–object distances and object–wall distances in realistic indoor environments. Importantly, the priors are not extracted from the 500 test scenes. Instead, we estimate the priors using an additional set of 200 non-Manhattan scenes processed with the same refinement pipeline, combined with 100 manually refined scenes from conventional Manhattan environments. We adopt this mixed prior estimation strategy because relying solely on non-Manhattan data may reduce the model’s generalization ability. The total number of samples used for prior estimation is limited to 300 due to the substantial human cost of data refinement: even with a semi-automatic annotation workflow, polishing each scene still requires approximately 10–15 minutes of annotator effort. Therefore, we only use 300 high-quality scenes for prior extraction, which we believe best reflect real-world object arrangement patterns. As shown in Tab.~\ref{tab:gen}, our method maintains strong generalization performance and still significantly outperforms existing approaches without re-estimating the priors.

\section{Detailed Definitions of Evaluation Metrics}
\label{supsec:1}

In this section, we provide the formal definitions and calculation details for the metrics used to evaluate our model. Let $\mathcal{S}$ denote a generated scene containing a set of $N$ objects, denoted as $\mathcal{O} = \{o_1, o_2, \dots, o_N\}$. Let $\mathcal{R}$ represent the 3D space occupied by the room boundary.

\subsection{Layout Violation Metrics}
To strictly evaluate the physical plausibility of the synthesized scenes, we employ mesh-based voxelization to calculate violations with high precision, rather than relying on coarse bounding box approximations. We report the following three metrics:

\begin{itemize}
    \item \textbf{Out-Of-Bounds (OOB):} This metric quantifies the volume of objects penetrating the room layout boundaries (e.g., floor, walls, ceiling). For each object $o_i$, we compute its voxelized mesh representation $\mathcal{V}(o_i)$. The OOB loss is defined as the total volume of object voxels lying outside the valid room space $\mathcal{R}$:
    \begin{equation}
        \mathcal{L}_{\text{OOB}} = \sum_{i=1}^{N} \text{Vol}(\mathcal{V}(o_i) \setminus \mathcal{R}),
    \end{equation}
    where $\text{Vol}(\cdot)$ denotes the physical volume calculated from the voxel occupancy, and $\setminus$ represents the set difference operation.

    \item \textbf{Mesh-Based Layout Loss (MBL):} This metric measures the severity of inter-object collisions. It is calculated as the sum of the intersection volumes between all pairs of distinct objects in the scene. Unlike bounding-box IoU, this metric accounts for the detailed geometry of the assets:
    \begin{equation}
        \mathcal{L}_{\text{MBL}} = \sum_{1 \le i < j \le N} \text{Vol}(\mathcal{V}(o_i) \cap \mathcal{V}(o_j)),
    \end{equation}
    where $\cap$ denotes the intersection of the voxelized meshes.

    \item \textbf{Voxelization-Based Loss (VBL):} VBL aggregates both boundary violations and inter-object collisions into a single measure of layout infeasibility:
    \begin{equation}
        \mathcal{L}_{\text{VBL}} = \mathcal{L}_{\text{OOB}} + \mathcal{L}_{\text{MBL}}.
    \end{equation}
    In our reported tables, these metrics ($\mathcal{L}_{\text{OOB}}$, $\mathcal{L}_{\text{MBL}}$, $\mathcal{L}_{\text{VBL}}$) are typically scaled (e.g., $\times 10^3$) for better readability.
\end{itemize}

\subsection{Layout Fidelity Metrics}
To assess how well the generated layouts align with human design distributions and the input textual intent, we utilize the following four metrics:

\begin{itemize}
    \item \textbf{Attribute Match (ATR):} ATR evaluates the semantic accuracy of the generated object categories. For a generated object $o_i$ and its expected category from the prompt $c_i^{gt}$, the score is calculated as:
    \begin{equation}
        \text{ATR} = \frac{1}{N} \sum_{i=1}^{N} \mathbb{I}(\text{Cat}(o_i) = c_i^{gt}),
    \end{equation}
    where $\mathbb{I}(\cdot)$ is the indicator function, and $\text{Cat}(\cdot)$ returns the semantic category of the object.

    \item \textbf{Object-Architecture Relationship (OAR):} This metric assesses the consistency of object placement relative to the room structure (walls). For each object $o_i$ belonging to category $c_i$, we measure its Euclidean distance to the nearest wall, denoted as $d(o_i, \mathcal{L})$. Using the category-specific wall distance mean $\mu_{c_i}^{w}$ and standard deviation $\sigma_{c_i}^{w}$ derived from the human-refined dataset statistics, OAR is calculated as the ratio of objects that fall within the expected statistical range:
    \begin{equation}
        \text{OAR} = \frac{1}{N} \sum_{i=1}^{N} \mathbb{I} \left( | d(o_i, \mathcal{L}) - \mu_{c_i}^{w} | \le \sigma_{c_i}^{w} \right),
    \end{equation}
    where $\mathbb{I}(\cdot)$ is the indicator function. A higher OAR indicates that the spatial distribution of objects relative to the room boundaries mimics human design habits.

    \item \textbf{Object-Object Relationship (OOR):} This metric measures whether the pairwise spatial arrangements between objects follow the learned prior distribution of the dataset. For any pair of objects $(o_i, o_j)$ with semantic categories $(c_i, c_j)$, let $\text{dist}(o_i, o_j)$ be the distance between their centers. Let $\mathcal{P}$ be the set of all unique pairs in the scene, where $|\mathcal{P}| = \frac{N(N-1)}{2}$. The OOR score is defined as:
    \begin{equation}
\label{eq:oor}
\begin{aligned}
\mathrm{OOR}
&=
\frac{1}{|\mathcal{P}|}
\sum_{(o_i,o_j)\in\mathcal{P}}
\mathbb{I}\Big(
\\
&\quad
\big|
\mathrm{dist}(o_i,o_j)
-
\mu^{p}_{c_i,c_j}
\big|
\le
\sigma^{p}_{c_i,c_j}
\Big),
\end{aligned}
\end{equation}
\noindent where $\mu_{c_i, c_j}^{p}$ and $\sigma_{c_i, c_j}^{p}$ represent the mean and standard deviation of the pairwise distance for the category combination $(c_i, c_j)$ as recorded in the statistical priors. This metric quantifies the plausibility of functional groupings, such as the typical proximity between a bed and a nightstand.

    \item \textbf{Prompt-Matching Score (PMS):} PMS quantifies the instruction-following capability by measuring the recall of words from the input prompt found in the description of the generated 3D asset. For a given input prompt $p_i$ and the description $d_i$ of the generated asset, the score is defined as:
    \begin{equation}
        \text{PMS}(p_i, d_i) = \frac{1}{|p_i|} \sum_{w_j \in p_i} \mathbb{I}_{w_j \in d_i},
    \end{equation}
    
    \noindent where $w_j$ represents the $j$-th word in the prompt $p_i$, $|p_i|$ denotes the total word count of the prompt, and $\mathbb{I}_{condition}$ is the indicator function that equals 1 if the word $w_j$ is present in the asset description $d_i$, and 0 otherwise. 
    
\end{itemize}

\section{The Detailed Implementation of Spatial Prior Guidance}
\label{supsec:2}

In this section, we provide a detailed and reproducible description of how the \textbf{Spatial Prior Guidance (SPG)} module is implemented, including (i) how the statistical spatial priors are computed from human-refined scenes, and (ii) how these priors are instantiated as a computable reward to guide layout generation. 

\subsection{Overview and Notation}
Given a generated scene $\mathcal{S}$ containing $N$ objects $\mathcal{O}=\{o_1,\dots,o_N\}$ and a room layout boundary $\mathcal{L}$, SPG introduces two types of priors:
\begin{itemize}
    \item \textbf{Object-Architecture Prior:} category-dependent wall distance statistics $(\mu_c^{w}, \sigma_c^{w})$.
    \item \textbf{Object-Object Prior:} category-pair-dependent relative distance statistics $(\mu_{c_i,c_j}^{p}, \sigma_{c_i,c_j}^{p})$.
\end{itemize}
These priors are estimated \textbf{offline} from the human-refined dataset, which exhibits a distribution more aligned with real-world scenarios, and are subsequently used during training to provide a dense reward signal that biases the model toward spatially plausible configurations.

\subsection{Estimating Object-Architecture Priors}
We first estimate the object-architecture priors used for evaluating and guiding \textbf{Object-Architecture Relationship (OAR)}.
For each object $o_i$ with semantic category $c_i$, we compute its Euclidean distance to the nearest wall in the room layout $\mathcal{L}$:
\begin{equation}
    d(o_i,\mathcal{L}).
\end{equation}

We then collect all wall-distance samples for each category $c$:
\begin{equation}
    \mathcal{D}_{c}^{w} = \left\{ d(o_i,\mathcal{L}) \;\middle|\; \text{Cat}(o_i)=c \right\}.
\end{equation}
Finally, we compute the category-wise mean and standard deviation:
\begin{equation}
\label{eq:stat}
\begin{aligned}
\mu_{c}^{w}
&=
\frac{1}{|\mathcal{D}_{c}^{w}|}
\sum_{d \in \mathcal{D}_{c}^{w}} d,
\\[4pt]
\sigma_{c}^{w}
&=
\sqrt{
\frac{1}{|\mathcal{D}_{c}^{w}|-1}
\sum_{d \in \mathcal{D}_{c}^{w}}
(d-\mu_{c}^{w})^2
}.
\end{aligned}
\end{equation}
These statistics capture human design preferences regarding object placements relative to architectural boundaries (e.g., beds and cabinets are often placed close to walls).

\subsection{Estimating Object-Object Priors}
We next estimate the object-object priors used for evaluating and guiding \textbf{Object-Object Relationship (OOR)}.
For any object pair $(o_i,o_j)$ with categories $(c_i,c_j)$, we compute their pairwise center distance:
\begin{equation}
    \text{dist}(o_i,o_j).
\end{equation}

We aggregate distances for each category pair $(c_i,c_j)$:
\begin{equation}
\label{eq:pairdist}
\begin{aligned}
\mathcal{D}_{c_i,c_j}^{p}
=
\Big\{
\mathrm{dist}(o_i,o_j)
\;\Big|\;
&\mathrm{Cat}(o_i)=c_i, \\
&\mathrm{Cat}(o_j)=c_j,\ i<j
\Big\}.
\end{aligned}
\end{equation}
Then, we compute the mean and standard deviation:
\begin{equation}
\label{eq:pair_stat}
\begin{aligned}
\mu_{c_i,c_j}^{p}
&=
\frac{1}{|\mathcal{D}_{c_i,c_j}^{p}|}
\sum_{d \in \mathcal{D}_{c_i,c_j}^{p}} d,
\\[4pt]
\sigma_{c_i,c_j}^{p}
&=
\sqrt{
\frac{1}{|\mathcal{D}_{c_i,c_j}^{p}|-1}
\sum_{d \in \mathcal{D}_{c_i,c_j}^{p}}
(d-\mu_{c_i,c_j}^{p})^2
}.
\end{aligned}
\end{equation}
These statistics reflect typical functional groupings in indoor scenes, such as the expected proximity between a bed and a nightstand.

\subsection{Instantiating $B(\cdot)$ and $A(\cdot)$ in SPG}

Given a scene $\mathcal{S}$ and an object $o_i$ with category $c_i$, SPG introduces two components modeled as Gaussian kernels:
\paragraph{Object-Architecture Term $B(\cdot)$.}The term $B(\cdot)$ quantifies the consistency of an object's placement with the category-specific wall distance distribution. It is formulated as a Gaussian function:\begin{equation}\label{eq:spg_B}B(o_i,\mathcal{L})= \exp \left( - \frac{(d(o_i, \mathcal{L}) - \mu_{c_i}^{w})^2}{2(\sigma_{c_i}^{w})^2} \right).\end{equation}

\paragraph{Object-Object Term $A(\cdot)$.}The term $A(\cdot)$ measures how well the relative distance between two objects conforms to the learned category-pair distribution:\begin{equation}\label{eq:spg_A}A(o_i,o_j)= \exp \left( - \frac{(\text{dist}(o_i, o_j) - \mu_{c_i, c_j}^{p})^2}{2(\sigma_{c_i, c_j}^{p})^2} \right).\end{equation}

\subsection{Discussion: Why SPG Improves OOR/OAR and Alleviates Collisions}
\paragraph{Improving OOR/OAR.}
By construction, SPG directly optimizes the same statistical consistency criteria used by OOR and OAR.
Therefore, maximizing $\mathcal{R}_{\text{SPG}}$ encourages the model to generate layouts that lie within the high-probability regions of human-refined spatial distributions, leading to higher OOR and OAR scores.

\paragraph{Alleviating Collisions.}
Although SPG does not explicitly compute mesh intersections, it implicitly discourages object collisions through the object-object prior.
In particular, when two objects overlap, their pairwise distance $\text{dist}(o_i,o_j)$ tends to be abnormally small and thus deviates from the learned mean $\mu_{c_i,c_j}^{p}$ for most functional category pairs.
As a result, such configurations are assigned low $\mathcal{R}_{\text{spg}}$, making them unlikely under the SPG-guided generation distribution.
This explains why SPG can reduce collision frequency in practice, even without explicitly using voxel-based collision penalties.
We emphasize that SPG serves as a \emph{statistical} plausibility prior rather than a hard feasibility constraint.
Thus, while SPG improves spatial rationality and reduces violations, strict collision-free guarantees require complementary geometric constraints, which are addressed by other components of the full framework.

\section{Expansion Capability of Our framework}\label{supsec:3}
Leveraging Structured Scene Representation (SSR), our framework goes beyond scene synthesis and naturally supports object-level scene editing, including removal and replacement of existing entities. Under the explicit, object-centric SSR formulation, object removal becomes a straightforward and interpretable procedure. Specifically, we first enumerate and inspect the set of instantiated objects in the current scene. We then employ a zero-shot large language model (LLM) to parse the user instruction and identify the target object category or entity to be removed. Next, we perform semantic matching between the extracted target description and the existing object candidates using CLIP-based similarity, and eliminate the most semantically aligned object from the SSR. Because subsequent 3D asset retrieval and scene rendering are conditioned on the updated SSR, the removed object is consistently absent from the regenerated scene.

\begin{figure}
    \centering
    \includegraphics[width=0.99\linewidth]{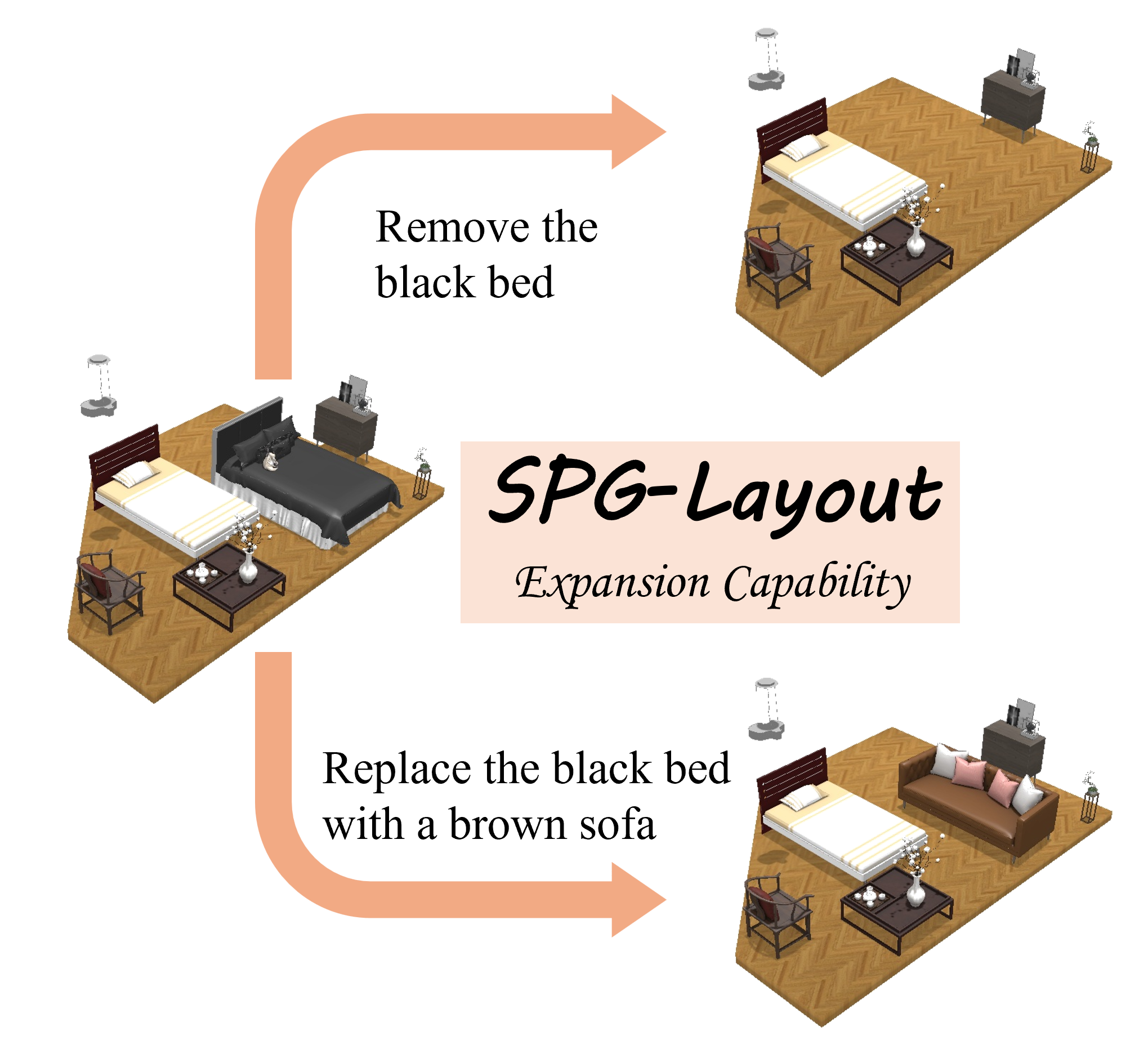}
    \caption{\textbf{Expansion Capability of SPG-Layout.}}
    \label{supfig:remove}
\end{figure}

Similarly, object replacement can be formulated as a composition of removal followed by insertion: we first remove the original object using the above strategy, and then introduce the requested new object into the SSR, triggering the corresponding asset retrieval and rendering steps. Fig.~\ref{supfig:remove} provides qualitative demonstrations of our framework on both remove and replace operations.

\section{3D Asset Sampling}\label{supsec:4}

We retrieve 3D assets using a stochastic sampling process that balances semantic alignment and geometric fit. For a generated object $o_i$ with description $d_i$ and target scale $s_i$, we assign a selection score to each candidate mesh $m_j$ in the asset library. The scoring function is defined as:
\begin{equation}
\label{eq:score}
\begin{aligned}
\mathrm{Score}(m_j)
&=
\lambda \cdot
\mathrm{sim}_{\mathrm{sem}}(d_i,d_j)
\\
&\quad
+
(1-\lambda)\cdot
\mathrm{sim}_{\mathrm{geo}}(s_i,s_j).
\end{aligned}
\end{equation}
where $d_j$ and $s_j$ denote the textual description and spatial dimensions of the candidate mesh $m_j$, respectively.

Specifically, the semantic similarity $\text{sim}_{\text{sem}}$ is computed via the dot product of L2-normalized SigLIP embeddings. The geometric similarity $\text{sim}_{\text{geo}}$ is calculated using a Gaussian kernel based on the Euclidean distance of 3D bounding box dimensions:
\begin{equation}
    \text{sim}_{\text{geo}}(s_i, s_j) = \exp \left( - \frac{\| s_i - s_j \|_2^2}{2\sigma^2} \right),
\end{equation}
where $\| \cdot \|_2$ denotes the $L_2$ norm. The final asset is selected via nucleus (top-$p$) sampling over the temperature-scaled softmax distribution derived from these scores. In line with the reference implementation, we set the weighting parameter $\lambda = 0.5$ to give equal importance to text description and spatial dimensions during retrieval.

\section{Additional Experimental Results}
\subsection{Independent Evaluation with SceneEval Metrics}\label{supsec:5}

To address potential concerns about reward-metric circularity, we additionally evaluate our method using the independent metrics from SceneEval~\cite{tam2024sceneeval}, which defines evaluation criteria independently of our reward design. As shown in Tab.~\ref{tab:sceneeval}, SPG-Layout consistently outperforms baselines across all independent metrics, confirming that our improvements reflect genuine layout quality rather than overfitting to the reward formulation.

\begin{table}[h]
\centering
\caption{\textbf{Independent evaluation using SceneEval metrics} on Full Scene Synthesis (non-Manhattan, `all' split).}
\label{tab:sceneeval}
\renewcommand{\arraystretch}{1.15}
\setlength{\tabcolsep}{3pt}
\resizebox{\linewidth}{!}{%
\begin{tabular}{l c c c c c c}
\toprule
\textbf{Method} & \textbf{CNT$\uparrow$} & \textbf{ATR$\uparrow$} & \textbf{COL$_{ob}$$\downarrow$} & \textbf{SUP$\uparrow$} & \textbf{NAV$\uparrow$} & \textbf{OOB$\downarrow$} \\
\midrule
ATISS & 10.72 & 7.29 & 56.27 & 82.77 & 78.12 & 31.86 \\
MiDiffusion & 12.85 & 11.32 & 43.78 & 88.58 & 88.96 & 18.97 \\
ReSpace & 65.81 & 38.49 & 2.37 & 98.32 & 98.71 & 0.16 \\
\rowcolor{gray!20} \textbf{SPG-Layout} & \textbf{82.33} & \textbf{55.76} & \textbf{0.24} & \textbf{99.23} & \textbf{99.40} & \textbf{0.03} \\
\bottomrule
\end{tabular}%
}
\end{table}

\subsection{Failure Cases}
\begin{figure}
    \centering
    \includegraphics[width=0.99\linewidth]{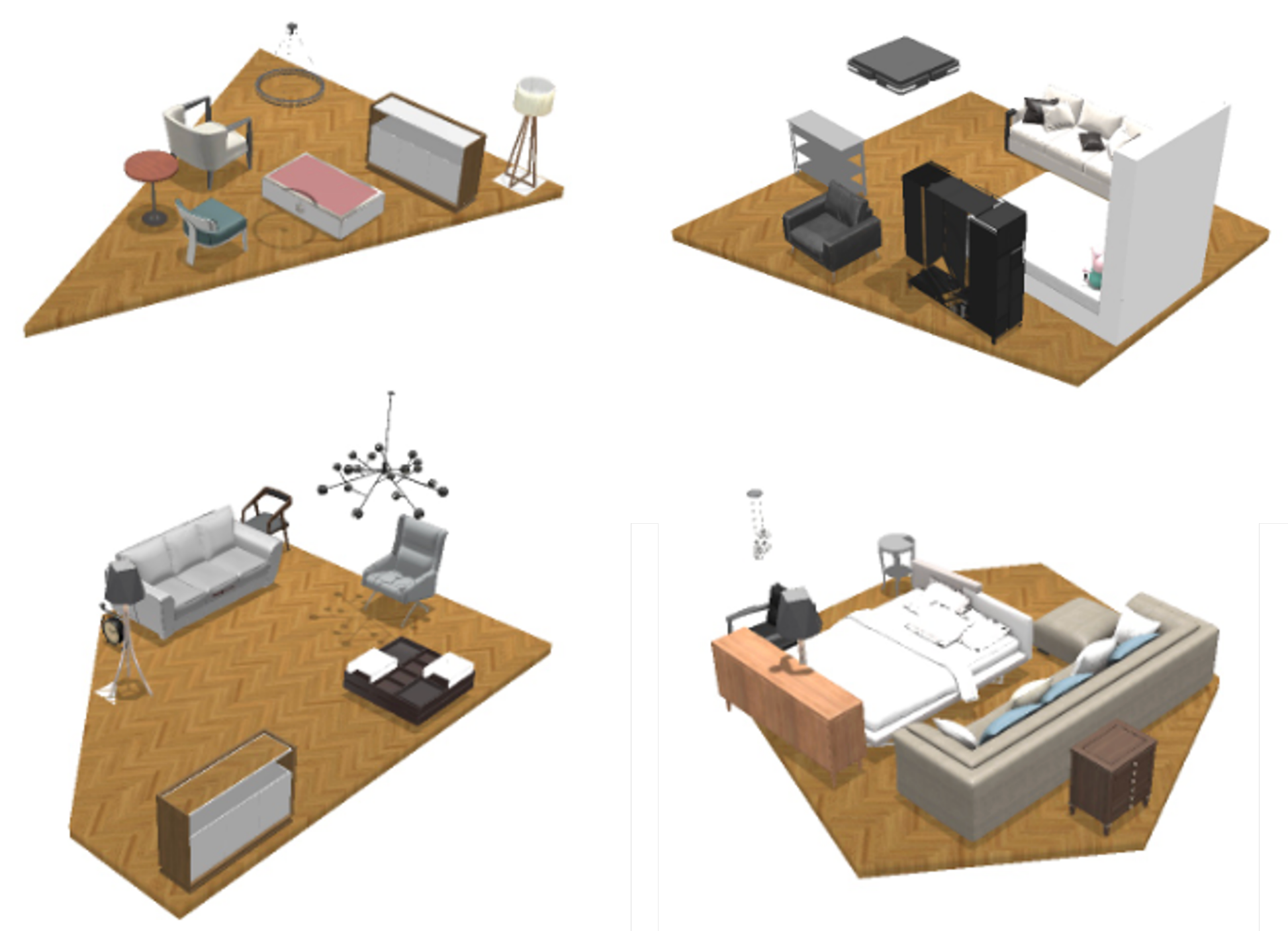}
    \caption{\textbf{Failure Cases.} Although no collisions occur, the generated layout severely deviates from human aesthetics, with cluttered and disorganized furniture arrangements.
}
    \label{fig:failurecase}
\end{figure}

Collisions and out-of-bound placements are very rare in SPG-Layout and generally occur only when scenes contain an excessive number of objects. The majority of current failure cases relate to functional failures, particularly the orientation of small objects. We hypothesize that large objects typically have more distinctive features and stronger spatial relationships with other objects, making their distributions easier for the network to capture. In contrast, small objects are more numerous and span more diverse categories, making it harder for the model to learn sufficiently discriminative features. Failure case visualizations are available at Fig.~\ref{fig:failurecase}.

\subsection{Additional Qualitative Results}
\begin{figure*}
    \centering
    \includegraphics[width=0.8\linewidth]{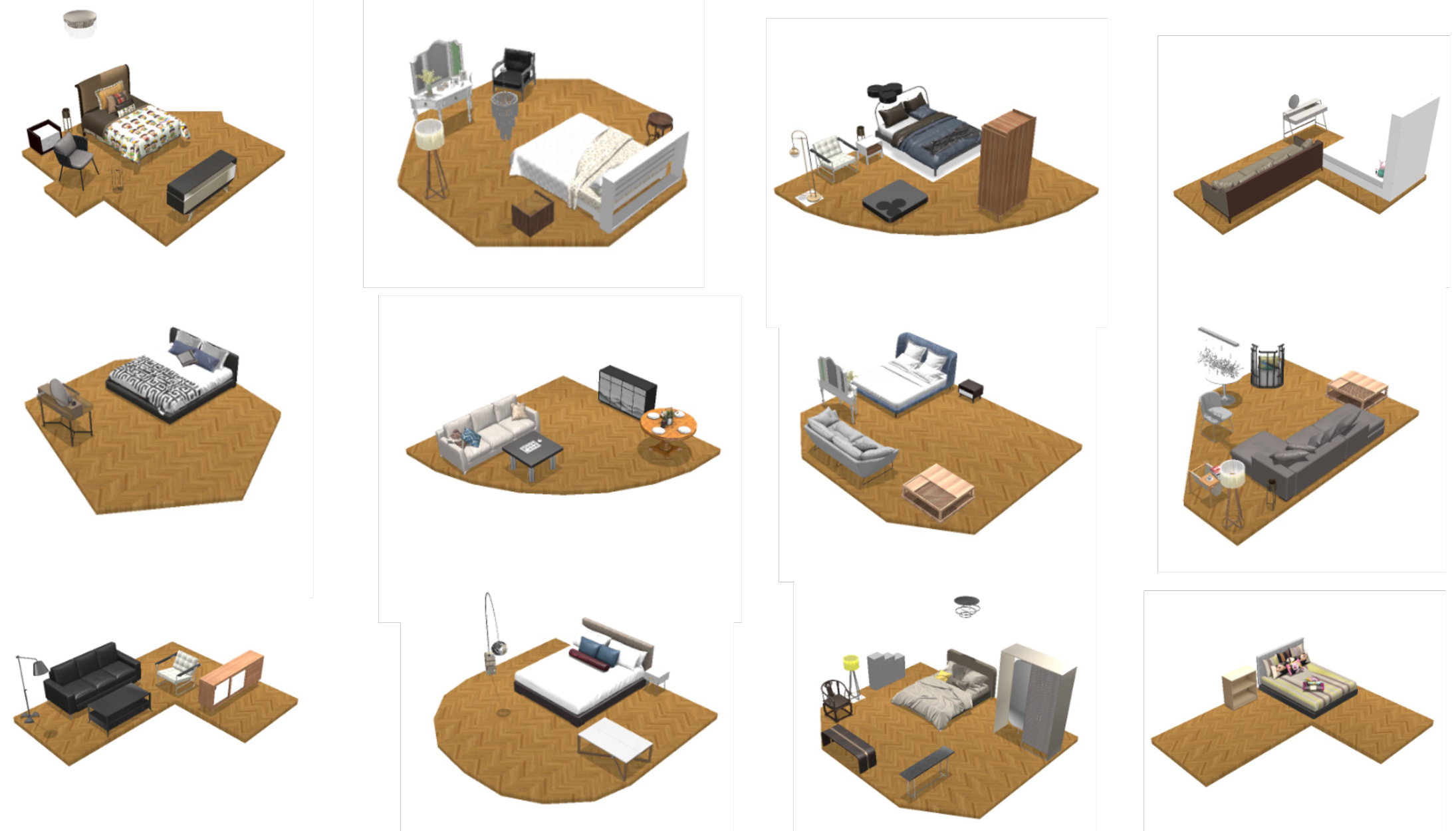}
    \caption{More Visualizations.}
    \label{fig:visresults}
\end{figure*}

\begin{figure*}
    \centering
    \includegraphics[width=0.6\linewidth]{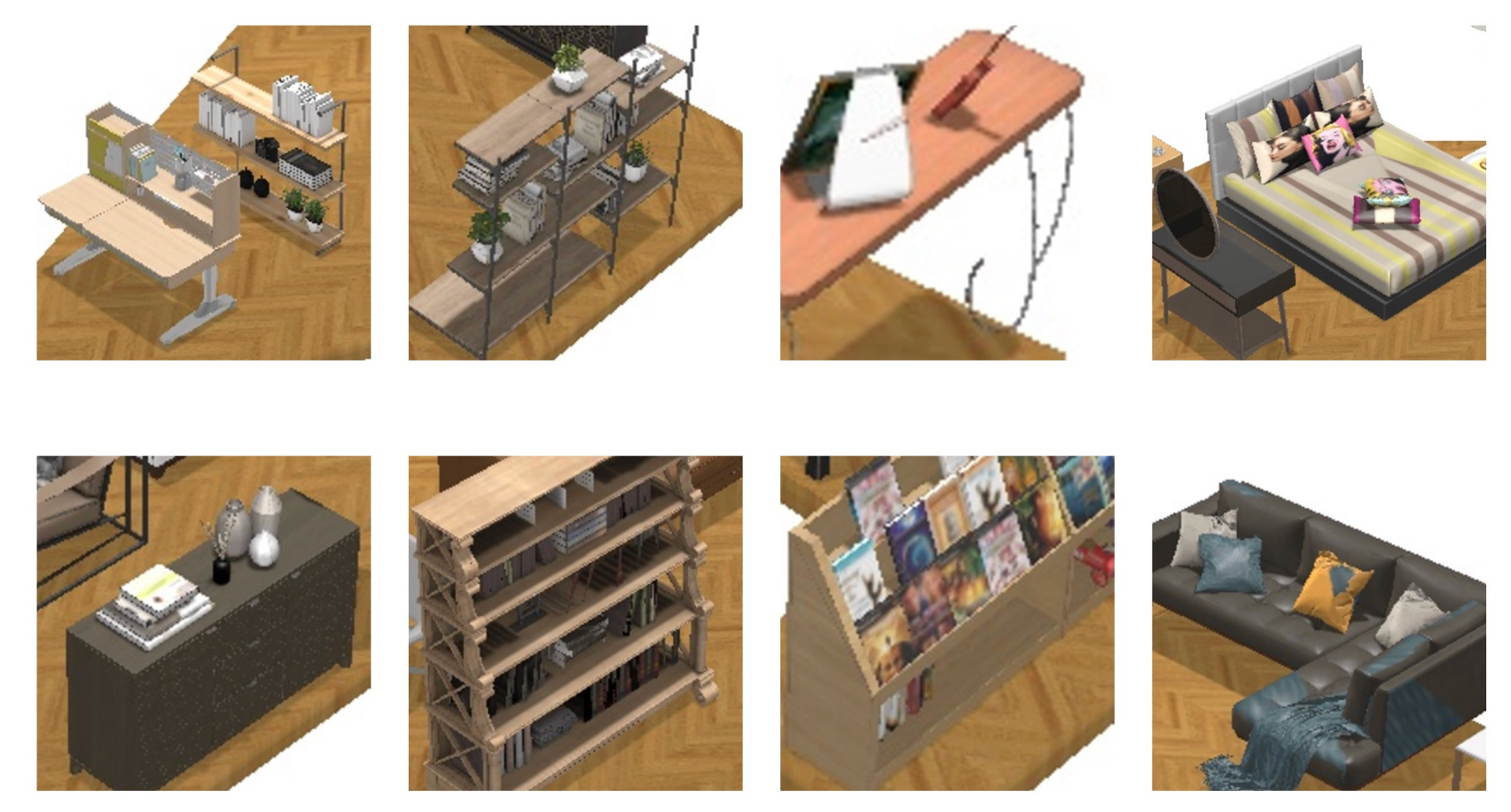}
    \caption{Stacked Cases.}
    \label{fig:stacked}
\end{figure*}

More visualization results are provided in this section, including additional visualization results generated by SPG-Layout (Fig.~\ref{fig:visresults}), as well as stacked scenes shown in Fig.~\ref{fig:stacked}.


\subsection{Additional Ablation Study}
To explore the impact of different aggregation operators in the SPG module, we conduct ablation experiments on the full scene synthesis task. The quantitative results are summarized in Table \ref{tab:ablation_agg}. We take two widely used aggregation strategies, Mean and Max, as baselines, and compare them with our proposed attention-based aggregation operator. Two evaluation metrics are adopted in our work: avg\_LV (lower is better) and avg\_LF (higher is better). We report the performance on bedroom (bed), living room (liv) and all scenes separately.

As shown in the table, Max aggregation achieves marginal improvements over Mean aggregation on most metrics, while it obtains slightly worse avg\_LV results on the bedroom scene. However, both simple aggregation methods cannot fully model complex feature interactions. By contrast, our attention-based aggregation achieves competitive performance and obtains the optimal results on liv and all scenes in terms of avg\_LV. Specifically, our method achieves avg\_LV values of 39.79, 6.74 and 19.52 for bed, liv and all scenes, respectively, reaching the lowest avg\_LV on living room and overall scenes. Meanwhile, our approach obtains the highest avg\_LF score of 77.14 on the living room, with avg\_LF scores of 77.27 and 78.36 for bedroom and all scenes correspondingly.

The above results indicate that static aggregation operations like Mean and Max fail to capture the importance of different features. Benefiting from adaptive weight assignment, the attention mechanism can effectively fuse multi-source features and model latent dependencies. This clearly verifies the rationality and effectiveness of our designed aggregation operator for full scene synthesis.

\begin{table}[t]
\centering
\caption{\textbf{Ablation of aggregation operators in SPG (Full Scene Synthesis).}}
\label{tab:ablation_agg}
\renewcommand{\arraystretch}{1.3}
\setlength{\tabcolsep}{4pt}

\resizebox{\linewidth}{!}{%
\begin{tabular}{l ccc ccc}
\toprule
\multirow{2}{*}{\textbf{Aggregation}} & \multicolumn{3}{c}{\textbf{avg\_LV} ($\downarrow$)} & \multicolumn{3}{c}{\textbf{avg\_LF} ($\uparrow$)} \\
\cmidrule(lr){2-4} \cmidrule(lr){5-7}
& \textbf{bed} & \textbf{liv} & \textbf{all} & \textbf{bed} & \textbf{liv} & \textbf{all} \\
\midrule

Mean   & 35.62 & 9.35 & 20.32 & 79.34 & 76.21 & 78.11 \\
Max    & 35.77 & 8.63 & 19.82 & 79.89 & 76.36 & 78.62 \\
\rowcolor{gray!20} \textbf{Attention (Ours)} & \textbf{39.79} & \textbf{6.74} & \textbf{19.52} & \textbf{77.27} & \textbf{77.14} & \textbf{78.36}   \\

\bottomrule
\end{tabular}%
}
\vspace{-3mm}
\end{table}

\section*{Limitations}
Although SPG-Layout achieves strong performance across diverse indoor environments, it still has several limitations. Its reliance on spatial priors learned from 300 scenes may limit generalization to long-tail object categories. During single-object editing, HLS may introduce unintended global rearrangements when no collision-free insertion exists. In addition, fine-grained orientation remains challenging, especially for asymmetric or small objects, leading to occasional functional failures. Finally, the voxel-based geometric reward improves physical plausibility but incurs substantial computational overhead, reducing training efficiency.

\bibliography{aaai2027}

@string{CVPR="Proceedings of the IEEE Conference on Computer Vision and Pattern Recognition (CVPR)"}

@string{ECCV="Proceedings of the European Conference on Computer Vision (ECCV)"}

@string{ICLR="Proceedings of the International Conference on Learning Representations (ICLR)"}

@string{ICML="Proceedings of the International Conference on Machine Learning (ICML)"}

@string{TOG="ACM Transactions on Graphics (TOG)"}

@string{CHI="ACM Conference on Human Factors in Computing Systems (CHI)"}

@string{WACV="Proc. of the Winter Conference on Applications of Computer Vision (WACV)"}

@inproceedings{wang2021sceneformer,
  title={Sceneformer: Indoor scene generation with transformers},
  author={Wang, Xinpeng and Yeshwanth, Chandan and Nie{\ss}ner, Matthias},
  booktitle={2021 International Conference on 3D Vision (3DV)},
  pages={106--115},
  year={2021},
  organization={IEEE}
}

@article{paschalidou2021atiss,
  title={Atiss: Autoregressive transformers for indoor scene synthesis},
  author={Paschalidou, Despoina and Kar, Amlan and Shugrina, Maria and Kreis, Karsten and Geiger, Andreas and Fidler, Sanja},
  journal={Advances in Neural Information Processing Systems},
  volume={34},
  pages={12013--12026},
  year={2021}
}

@article{hu2024mixed,
  title={Mixed Diffusion for 3D Indoor Scene Synthesis},
  author={Hu, Siyi and Arroyo, Diego Martin and Debats, Stephanie and Manhardt, Fabian and Carlone, Luca and Tombari, Federico},
  journal={arXiv preprint arXiv:2405.21066},
  year={2024}
}

@inproceedings{fu20213d_front,
  title={3d-front: 3d furnished rooms with layouts and semantics},
  author={Fu, Huan and Cai, Bowen and Gao, Lin and Zhang, Ling-Xiao and Wang, Jiaming and Li, Cao and Zeng, Qixun and Sun, Chengyue and Jia, Rongfei and Zhao, Binqiang and others},
  booktitle={Proceedings of the IEEE/CVF International Conference on Computer Vision},
  pages={10933--10942},
  year={2021}
}

@article{lin2024instructscene,
  title={Instructscene: Instruction-driven 3d indoor scene synthesis with semantic graph prior},
  author={Lin, Chenguo and Mu, Yadong},
  journal={arXiv preprint arXiv:2402.04717},
  year={2024}
}

@inproceedings{kalervo2019cubicasa5k,
  author    = {Ahti Kalervo and Juha Ylioinas and Markus H{\"a}iki{\"o} and Antti Karhu and Juho Kannala},
  title     = {CubiCasa5K: A Dataset and an Improved Multi-Task Model for Floorplan Image Analysis},
  booktitle = {Proceedings of the IEEE/CVF International Conference on Computer Vision Workshops (ICCVW)},
  year      = {2019}
}

@inproceedings{sun2019horizonnet,
  author    = {Cheng Sun and Chi{-}Wei Hsiao and Min Sun and Hwann{-}Tzong Chen},
  title     = {HorizonNet: Learning Room Layout with 1D Representation and Pano Stretch Data Augmentation},
  booktitle = {Proceedings of the IEEE/CVF Conference on Computer Vision and Pattern Recognition (CVPR)},
  year      = {2019},
  pages     = {1047--1056}
}

@inproceedings{tam2026sceneeval,
  author    = {Hou In Ivan Tam and
               Hou In Derek Pun and
               Austin T. Wang and
               Angel X. Chang and
               Manolis Savva},
  title     = {SceneEval: Evaluating Semantic Coherence in Text-Conditioned 3D Indoor Scene Synthesis},
  booktitle = {Proceedings of the IEEE/CVF Winter Conference on Applications of Computer Vision (WACV)},
  year      = {2026}
}

@inproceedings{zou2018layoutnet,
  author    = {Chuhang Zou and Alex Colburn and Qi Shan and Derek Hoiem},
  title     = {LayoutNet: Reconstructing the 3D Room Layout from a Single RGB Image},
  booktitle = {Proceedings of the IEEE/CVF Conference on Computer Vision and Pattern Recognition (CVPR)},
  year      = {2018},
  pages     = {2051--2059}
}

@article{shao2024deepseekmath,
  title={Deepseekmath: Pushing the limits of mathematical reasoning in open language models},
  author={Shao, Zhihong and Wang, Peiyi and Zhu, Qihao and Xu, Runxin and Song, Junxiao and Bi, Xiao and Zhang, Haowei and Zhang, Mingchuan and Li, YK and Wu, Y and others},
  journal={arXiv preprint arXiv:2402.03300},
  year={2024}
}

@inproceedings{celen2024idesign,
  author    = {Ata Çelen and Guo Han and Konrad Schindler and Luc Van Gool and Iro Armeni and Anton Obukhov and Xi Wang},
  title     = {I-Design: Personalized LLM Interior Designer},
  booktitle = {Computer Vision -- ECCV 2024 Workshops},
  series     = {Lecture Notes in Computer Science (LNCS)},
  publisher  = {Springer},
  year       = {2025},
  pages      = {217--234},
  doi        = {10.1007/978-3-031-92387-6_17}
}

@inproceedings{fang2025ctrlroom,
  author    = {Chuan Fang and Yuan Dong and Kunming Luo and Xiaotao Hu and Rakesh Shrestha and Ping Tan},
  title     = {Ctrl-Room: Controllable Text-to-3D Room Meshes Generation with Layout Constraints},
  booktitle = {Proceedings of the International Conference on 3D Vision (3DV)},
  year      = {2025},
  pages     = {692--701},
  publisher = {IEEE},
  doi       = {10.1109/3DV66043.2025.00069}
}

@article{weiss2018,
  title={Fast and scalable position-based layout synthesis},
  author={Weiss, Tomer and Litteneker, Alan and Duncan, Noah and Nakada, Masaki and Jiang, Chenfanfu and Yu, Lap-Fai and Terzopoulos, Demetri},
  journal={IEEE Transactions on Visualization and Computer Graphics},
  volume={25},
  number={12},
  pages={3231--3243},
  year={2018},
  publisher={IEEE}
}

@inproceedings{qi2018,
  author    = {Siyuan Qi and Yixin Zhu and Siyuan Huang and Chenfanfu Jiang and Song{-}Chun Zhu},
  title     = {Human-Centric Indoor Scene Synthesis Using Stochastic Grammar},
  booktitle = {Proceedings of the IEEE/CVF Conference on Computer Vision and Pattern Recognition (CVPR)},
  year      = {2018},
  pages     = {5899--5908}
}

@article{fisher2015,
  author  = {Matthew Fisher and Manolis Savva and Pat Hanrahan and Matthias Nie{\ss}ner},
  title   = {Activity-Centric Scene Synthesis for Functional 3D Scene Modeling},
  journal = {ACM Transactions on Graphics (TOG)},
  volume  = {34},
  number  = {6},
  pages   = {179:1--179:13},
  year    = {2015}
}

@article{purkait2020,
  author  = {Pulak Purkait and Christopher Zach and Tat-Jun Chin and Ian Reid},
  title   = {Flexible Indoor Scene Synthesis Based on Multi-object Particle Swarm Intelligence Optimization and User Intentions with 3D Gesture},
  journal = {Computers \& Graphics},
  volume  = {93},
  pages   = {1--12},
  year    = {2020}
}

@inproceedings{ritchie2019fast,
  title={Fast and flexible indoor scene synthesis via deep convolutional generative models},
  author={Ritchie, Daniel and Wang, Kai and Lin, Yu-an},
  booktitle={Proceedings of the IEEE/CVF conference on computer vision and pattern recognition},
  pages={6182--6190},
  year={2019}
}

@inproceedings{tang2024diffuscene,
  title={Diffuscene: Denoising diffusion models for generative indoor scene synthesis},
  author={Tang, Jiapeng and Nie, Yinyu and Markhasin, Lev and Dai, Angela and Thies, Justus and Nie{\ss}ner, Matthias},
  booktitle={Proceedings of the IEEE/CVF conference on computer vision and pattern recognition},
  pages={20507--20518},
  year={2024}
}

@inproceedings{wei2023lego,
  title={Lego-net: Learning regular rearrangements of objects in rooms},
  author={Wei, Qiuhong Anna and Ding, Sijie and Park, Jeong Joon and Sajnani, Rahul and Poulenard, Adrien and Sridhar, Srinath and Guibas, Leonidas},
  booktitle={Proceedings of the IEEE/CVF Conference on Computer Vision and Pattern Recognition},
  pages={19037--19047},
  year={2023}
}

@article{wang2018deep,
  title={Deep convolutional priors for indoor scene synthesis},
  author={Wang, Kai and Savva, Manolis and Chang, Angel X and Ritchie, Daniel},
  journal={ACM Transactions on Graphics (TOG)},
  volume={37},
  number={4},
  pages={1--14},
  year={2018},
  publisher={ACM New York, NY, USA}
}

@inproceedings{zhai2024echoscene,
  title={Echoscene: Indoor scene generation via information echo over scene graph diffusion},
  author={Zhai, Guangyao and {\"O}rnek, Evin P{\i}nar and Chen, Dave Zhenyu and Liao, Ruotong and Di, Yan and Navab, Nassir and Tombari, Federico and Busam, Benjamin},
  booktitle={European Conference on Computer Vision},
  pages={167--184},
  year={2024},
  organization={Springer}
}

@article{maillard2024debara,
  title={Debara: Denoising-based 3d room arrangement generation},
  author={Maillard, L{\'e}opold and Sereyjol-Garros, Nicolas and Durand, Tom and Ovsjanikov, Maks},
  journal={Advances in Neural Information Processing Systems},
  volume={37},
  pages={109202--109232},
  year={2024}
}

@article{feng2023layoutgpt,
  title={Layoutgpt: Compositional visual planning and generation with large language models},
  author={Feng, Weixi and Zhu, Wanrong and Fu, Tsu-jui and Jampani, Varun and Akula, Arjun and He, Xuehai and Basu, Sugato and Wang, Xin Eric and Wang, William Yang},
  journal={Advances in Neural Information Processing Systems},
  volume={36},
  pages={18225--18250},
  year={2023}
}

@inproceedings{yang2024holodeck,
  title={Holodeck: Language guided generation of 3d embodied ai environments},
  author={Yang, Yue and Sun, Fan-Yun and Weihs, Luca and VanderBilt, Eli and Herrasti, Alvaro and Han, Winson and Wu, Jiajun and Haber, Nick and Krishna, Ranjay and Liu, Lingjie and others},
  booktitle={Proceedings of the IEEE/CVF Conference on Computer Vision and Pattern Recognition},
  pages={16227--16237},
  year={2024}
}

@article{sun2024layoutvlm,
  title={LayoutVLM: Differentiable Optimization of 3D Layout via Vision-Language Models},
  author={Sun, Fan-Yun and Liu, Weiyu and Gu, Siyi and Lim, Dylan and Bhat, Goutam and Tombari, Federico and Li, Manling and Haber, Nick and Wu, Jiajun},
  journal={arXiv preprint arXiv:2412.02193},
  year={2024}
}

@article{aguina2024open,
  title={Open-universe indoor scene generation using llm program synthesis and uncurated object databases},
  author={Aguina-Kang, Rio and Gumin, Maxim and Han, Do Heon and Morris, Stewart and Yoo, Seung Jean and Ganeshan, Aditya and Jones, R Kenny and Wei, Qiuhong Anna and Fu, Kailiang and Ritchie, Daniel},
  journal={arXiv preprint arXiv:2403.09675},
  year={2024}
}

@article{yang2024llplace,
  title={Llplace: The 3d indoor scene layout generation and editing via large language model},
  author={Yang, Yixuan and Lu, Junru and Zhao, Zixiang and Luo, Zhen and Yu, James JQ and Sanchez, Victor and Zheng, Feng},
  journal={arXiv preprint arXiv:2406.03866},
  year={2024}
}

@article{yang2024qwen2,
  title={Qwen2. 5 technical report},
  author={Yang, An and Yang, Baosong and Zhang, Beichen and Hui, Binyuan and Zheng, Bo and Yu, Bowen and Li, Chengyuan and Liu, Dayiheng and Huang, Fei and Wei, Haoran and others},
  journal={arXiv preprint arXiv:2412.15115},
  year={2024}
}

@article{hu2022lora,
  title={Lora: Low-rank adaptation of large language models.},
  author={Hu, Edward J and Shen, Yelong and Wallis, Phillip and Allen-Zhu, Zeyuan and Li, Yuanzhi and Wang, Shean and Wang, Lu and Chen, Weizhu and others},
  journal={ICLR},
  volume={1},
  number={2},
  pages={3},
  year={2022}
}

@inproceedings{yang2024physcene,
  title={Physcene: Physically interactable 3d scene synthesis for embodied ai},
  author={Yang, Yandan and Jia, Baoxiong and Zhi, Peiyuan and Huang, Siyuan},
  booktitle={Proceedings of the IEEE/CVF Conference on Computer Vision and Pattern Recognition},
  pages={16262--16272},
  year={2024}
}

@inproceedings{sun2024forest2seq,
  title={Forest2seq: Revitalizing order prior for sequential indoor scene synthesis},
  author={Sun, Qi and Zhou, Hang and Zhou, Wengang and Li, Li and Li, Houqiang},
  booktitle={European Conference on Computer Vision},
  pages={251--268},
  year={2024}
}

@inproceedings{feng2025casagpt,
  title={CasaGPT: cuboid arrangement and scene assembly for interior design},
  author={Feng, Weitao and Zhou, Hang and Liao, Jing and Cheng, Li and Zhou, Wenbo},
  booktitle={Proceedings of the Computer Vision and Pattern Recognition Conference},
  pages={29173--29182},
  year={2025}
}

@article{wang2023robogen,
  title={Robogen: Towards unleashing infinite data for automated robot learning via generative simulation},
  author={Wang, Yufei and Xian, Zhou and Chen, Feng and Wang, Tsun-Hsuan and Wang, Yian and Fragkiadaki, Katerina and Erickson, Zackory and Held, David and Gan, Chuang},
  journal={arXiv preprint arXiv:2311.01455},
  year={2023}
}

@inproceedings{zheng2020structured3d,
  title={Structured3d: A large photo-realistic dataset for structured 3d modeling},
  author={Zheng, Jia and Zhang, Junfei and Li, Jing and Tang, Rui and Gao, Shenghua and Zhou, Zihan},
  booktitle={European Conference on Computer Vision},
  pages={519--535},
  year={2020}
}

@inproceedings{song2025hazards,
  title={Hazards in daily life? enabling robots to proactively detect and resolve anomalies},
  author={Song, Zirui and Ouyang, Guangxian and Fang, Meng and Na, Hongbin and Shi, Zijing and Chen, Zhenhao and Yujie, Fu and Zhang, Zeyu and Jiang, Shiyu and Fang, Miao and others},
  booktitle={Proceedings of the 2025 Conference of the Nations of the Americas Chapter of the Association for Computational Linguistics: Human Language Technologies (Volume 1: Long Papers)},
  pages={7399--7415},
  year={2025}
}

@article{yang2025sceneweaver,
  title={SceneWeaver: All-in-One 3D Scene Synthesis with an Extensible and Self-Reflective Agent},
  author={Yang, Yandan and Jia, Baoxiong and Zhang, Shujie and Huang, Siyuan},
  journal={arXiv preprint arXiv:2509.20414},
  year={2025}
}

@article{bucher2025respace,
  title={ReSpace: Text-Driven 3D Scene Synthesis and Editing with Preference Alignment},
  author={Bucher, Martin JJ and Armeni, Iro},
  journal={arXiv preprint arXiv:2506.02459},
  year={2025}
}

@inproceedings{fu2025anyhome,
  title={{AnyHome}: Open-vocabulary generation of structured and textured {3D} homes},
  author={Fu, Rao and Wen, Zehao and Liu, Zichen and Sridhar, Srinath},
  booktitle=ECCV,
  pages={52--70},
  year={2024}
}

@inproceedings{hu2024scenecraft,
  title={{SceneCraft}: An {LLM} Agent for Synthesizing {3D} Scenes as {Blender} Code},
  author={Hu, Ziniu and Iscen, Ahmet and Jain, Aashi and Kipf, Thomas and Yue, Yisong and Ross, David A and Schmid, Cordelia and Fathi, Alireza},
  booktitle=ICML,
  year={2024},
  volume={235},
  pages={19252--19282}
}

@inproceedings{littlefair2025flairgpt,
  title={{FlairGPT}: Repurposing {LLMs} for interior designs},
  author={Littlefair, Gabrielle and Dutt, Niladri Shekhar and Mitra, Niloy J},
  booktitle={Computer Graphics Forum},
  pages={e70036},
  year={2025}
}

@article{wang2024chat2layout,
  title={{Chat2Layout: Interactive 3D Furniture Layout with a Multimodal LLM}},
  author={Can Wang and Hongliang Zhong and Menglei Chai and Mingming He and Dongdong Chen and Jing Liao},
  journal={IEEE transactions on visualization and computer graphics},
  year={2024},
  volume={PP}
}

@inproceedings{xia2024scenegenagent,
  title={{SceneGenAgent}: Precise Industrial Scene Generation with Coding Agent},
  author={Xia, Xiao and Zhang, Dan and Liao, Zibo and Hou, Zhenyu and Sun, Tianrui and Li, Jing and Fu, Ling and Dong, Yuxiao},
  booktitle={Proceedings of the 63rd Annual Meeting of the Association for Computational Linguistics},
  pages={17847--17875},
  year={2025}
}

@inproceedings{ccelen2025design,
  title={{I-Design}: Personalized {LLM} interior designer},
  author={{\c{C}}elen, Ata and Han, Guo and Schindler, Konrad and Van Gool, Luc and Armeni, Iro and Obukhov, Anton and Wang, Xi},
  booktitle=ECCV,
  pages={217--234},
  year={2024}
}

@article{hao2025mimo,
  title={MiMo-Embodied: X-Embodied Foundation Model Technical Report},
  author={Hao, Xiaoshuai and Zhou, Lei and Huang, Zhijian and Hou, Zhiwen and Tang, Yingbo and Zhang, Lingfeng and Li, Guang and Lu, Zheng and Ren, Shuhuai and Meng, Xianhui and others},
  journal={arXiv preprint arXiv:2511.16518},
  year={2025}
}

\end{document}